%% file: main.tex
\documentclass[lettersize,journal]{IEEEtran}

\usepackage{cite}
\usepackage{amsmath,amssymb,amsfonts}
\usepackage{algorithmic}
\usepackage{graphicx}
\usepackage{textcomp}
\usepackage{xcolor}
\usepackage{multirow}
\usepackage{graphicx}
\usepackage{caption}
\usepackage{subcaption}
\usepackage{booktabs}
\usepackage{bbm}
\usepackage{array}
\usepackage{makecell}
\usepackage{booktabs}
\usepackage{hyperref}
\usepackage{balance}

\begin{document}

\title{LLMApp: Unleashing the Power of Large Language Model for Mobile App Usage Prediction}

\author{Bo~Liu,
        Tong~Li,~\IEEEmembership{Member,~IEEE,}
        Miao~Xiao,
        Beihao~Xia, \\
        Zhu~Xiao,~\IEEEmembership{Senior Member,~IEEE,}
        Zhuo~Tang,
        Kenli~Li
        
\IEEEcompsocitemizethanks
{\IEEEcompsocthanksitem Manuscript received XX  XX 2025.~This work was supported in part by the National Natural Science Foundation of China under Grant U24A20247, Grant 62471277 and Grant 62671244, in part by the Young Elite Scientists Sponsorship Program by CAST under Grant ZB2025-293, in part by the Fundamental Research Funds for the Central Universities under Grant 531119200354, in part by the National Natural Science Foundation of
Hunan Province under Grant 2025JJ50351,~and in part by Shenzhen Science and Technology Program under Grant JCYJ20250604185802004.~\textit{(Corresponding authors:~Tong~Li and Zhu~Xiao.)}

B. Liu, T. Li, M. Xiao,~Z. Tang, and K. Li are with the College of Computer Science and Electronic Engineering, the Ministry of Education Key Laboratory of ``Fusion Computing of Supercomputing and Artificial Intelligence'', Hunan University, Changsha 410082, China.  E-mail:~\{liubo317; tliay; ztang; lkl\}@hnu.edu.cn.

Z. Xiao is with the College of Computer Science and Electronic Engineering, Hunan University, Changsha 410082, China, and also with Shenzhen Research Institute, Hunan University, Shenzhen 518055, China.  E-mail:~zhxiao@hnu.edu.cn.

B.~Xia is with the School of Electronic Information and Communications, Huazhong University of Science and Technology, Wuhan, China, and with the Department of Urban Planning and Design, The University of Hong Kong, 
Pokfulam Road, Hong Kong SAR, China. E-mail:~xbh\_hust@connect.hku.hk. 

}
}

\markboth{IEEE Transactions on Services Computing}%
{How to Use the IEEEtran \LaTeX \ Templates}

\maketitle

\begin{abstract}
Mobile applications serve as gateways to many essential services in human daily life. Mobile app usage prediction can help reduce service delay, support better resource allocation, and assist network operators in managing traffic changes, ultimately improving the quality and responsiveness of mobile services.
In this paper, we propose LLMApp, a hierarchical prediction framework that leverages large language models to predict mobile app usage. Specifically, LLMApp employs a prompt compressor to filter redundant long-term usage logs, producing concise summary prompts that the LLM can effectively reason over. A context interpreter encodes diverse contextual information into unified representations and projects them into the LLM's semantic space, enabling joint interpretation of spatiotemporal and behavioral signals. Moreover, a hierarchical predictor leverages LLMs' semantic reasoning capabilities to perform coarse-to-fine mobile app usage prediction.
Experiments on two real-world datasets show that LLMApp outperforms baseline approaches by over 9\% across Top-k Accuracy (ACC@K), Mean Reciprocal Rank (MRR@K), and Normalized Discounted Cumulative Gain (NDCG@K). Further analysis demonstrates that LLMApp exhibits strong robustness and scalability, effectively filters for redundancy, and accurately captures the spatiotemporal patterns underlying user app usage behavior. The code is available at: https://github.com/BoLiupro/LLMApp.

\end{abstract}


\input{Introduction}
\input{Preliminaries}
\input{Methods}

\input{Experiments}
\input{Related}
\section{Limitation}
\textcolor{black}{
One limitation of this study is that the experiments are conducted on two city-scale app usage datasets collected from Shanghai and Nanchang. Although these datasets contain large-scale real-world records and span different urban environments, they may not fully represent app usage behaviors across culturally distinct regions or among diverse user groups. This limitation mainly arises from the privacy-sensitive nature of mobile app usage data, since fine-grained app records are usually associated with users' spatiotemporal traces and behavioral preferences, thereby limiting the availability of publicly available large-scale datasets. Future work can further evaluate LLMApp on more diverse regions and user groups when suitable datasets become available. In addition, privacy-preserving data collection, federated learning, and synthetic app usage data generation can be explored to support broader evaluation while reducing the risk of exposing sensitive user information.
}
\input{Conclusion}


\balance
\bibliographystyle{IEEEtran}
\bibliography{Ref}

\end{document}

%% file: Introduction.tex
\section{Introduction}
\label{sec:Intro}

With the rapid development of mobile networks and digital service ecosystems~\cite{li2020apps,zhao2019appusage2vec}, mobile applications (apps) have become gateways to many essential mobile services in daily life, ranging from commerce~\cite{gruning2023directing} and entertainment~\cite{jabeur2013mobile} to transportation~\cite{chu2021air} and healthcare~\cite{munzert2021tracking}. As users increasingly rely on mobile apps, delays, unstable resource usage, or inefficient traffic handling can interrupt interactions and degrade the overall service experience~\cite{li2022smartphone}. An effective strategy to mitigate these issues is to predict users' app usage and preload likely-following apps into memory. Such prediction can reduce service delay, enable better resource allocation, and assist network operators in managing traffic dynamics, ultimately improving the quality and responsiveness of mobile services.

\begin{figure}[tb]
\vspace{-2mm}
\centering
\includegraphics[width=1.01\linewidth]{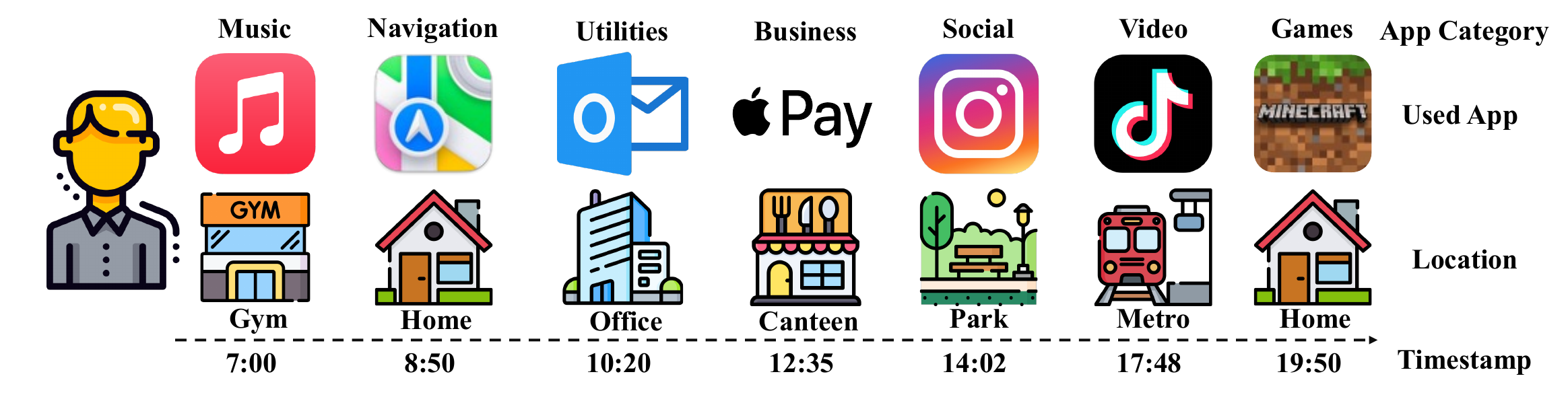}
\vspace{-3mm}
\caption{Spatiotemporal contexts shape mobile app usage of a user.}
\label{fig:app_usage}
\vspace{-4mm}
\end{figure}

Existing methods have devoted considerable effort to predicting app usage, progressing from short-term sequence models to context-augmented approaches and, ultimately, to structured or relational modeling. Early studies applied Markov models and recurrent networks to capture short-term transitions between consecutively used apps~\cite{shin2012understanding,zhao2019appusage2vec,sun2025appformer,zhang2023predicting}. Some studies further incorporated contextual signals such as location, time, and user profiles~\cite{liao2013mining,lee2019app,leroux2013mobile}, modeling app usage as a function of latent behavioral states, context transitions, or dynamic relational structures. More recent advances introduced richer structural modeling. For example, Sadhu~\emph{et al.}~\cite{sadhu2022deepcontext} jointly modeled mobile context evolution using deep neural networks, Shen~\emph{et al.}~\cite{shen2019deepapp} formulated app prediction as a reinforcement learning task, and Huang~\emph{et al.}~\cite{huang2025predicting} used hypergraphs to capture higher-order and temporal dependencies.

Despite remarkable progress, previous studies still face several limitations. One limitation is that existing models cannot effectively reason about the semantic relationships between real-world contexts and app usage. Most mobile app usage prediction models are numerical association models that rely on statistical co-occurrence patterns and sequential correlations. As a result, they lack semantic understanding and reasoning capabilities. For example, as illustrated in Fig.~\ref{fig:app_usage}, previous models fail to infer that a situation such as `the user has just finished a calendar event at the office, it is around 1 p.m., and he is in a canteen' implicitly suggests that it is lunchtime, during which payment apps are likely to be opened.
Another major limitation lies in the complicated and fragmented treatment of contextual information. Prior models typically depend on hand-crafted feature engineering or multi-branch embedding pipelines to incorporate diverse contextual signals. Such designs lack flexibility and overlook deeper cross-modal relationships that naturally occur in real-world scenarios, preventing existing methods from fully capturing the rich interactions among heterogeneous contextual signals.

Large Language Models (LLMs) have recently demonstrated remarkable proficiency across a wide range of natural language processing tasks. Their strengths in text understanding and semantic reasoning make them promising tools for domains beyond traditional language processing. For example, MAPLE~\cite{khaokaew2024maple} adopts a prompt-based framework that aligns app usage sequences and contextual information with natural-language semantics to predict the next app. The ability of LLMs to unify diverse contexts and perform semantic, context-aware reasoning helps address the limitations of previous app usage prediction models. It offers new opportunities for advancing mobile app usage prediction.

In this study, we aim to unleash the power of LLMs to predict mobile app usage. Achieving this objective requires addressing the following three challenges:
$(1)$ \textbf{How can LLMs understand and reason about the dependencies between spatiotemporal context and app usage semantics?} 
Users' mobile app usage often exhibits distinct patterns across different locations and times. Leveraging LLMs for app usage prediction requires the model to understand how these contextual factors jointly shape user behavior. However, conventional LLMs are fundamentally designed to model dependencies within textual token sequences. In contrast, the dependencies between spatiotemporal context and app usage semantics reflect high-order human behavioral logic that differs from textual patterns, making it challenging for LLMs to directly learn the underlying associations between spatiotemporal context and app usage semantics.
$(2)$ \textbf{How can long and noisy app usage histories be compressed into compact and informative summary prompts for LLMs?} 
Mobile app usage logs are typically short, frequent, and highly fragmented~\cite{bohmer2011falling}, resulting in long histories filled with redundant and weakly informative records. Directly converting these verbose logs into prompts and feeding them into an LLM not only greatly increases computational cost but also introduces noise that can obscure meaningful behavioral cues. Therefore, removing redundancy while preserving the most representative usage signals is crucial for enabling LLMs to infer user behavior efficiently.
$(3)$ \textbf{ How can diverse contextual signals be encoded into a unified semantic space interpretable by LLMs?} 
Contexts such as location, time, app category, and traffic level have heterogeneous formats and structural patterns, whereas LLMs operate on linguistic representations. Enabling LLMs to understand the semantics of these contextual signals requires encoding them into a continuous and unified embedding space. Such unified representations enable LLMs to interpret relationships among signals and perform joint reasoning across spatiotemporal context and app usage behavior.

To overcome the challenges outlined above, we propose LLMApp, an LLM-based hierarchical model for predicting mobile app usage. 
Given that spatiotemporal context significantly influences mobile app usage behavior~\cite{li2021understanding,yu2020semantic}, and users' usage patterns differ across app categories and traffic levels across locations and times, we decompose the prediction process into two phases. Based on strong contextual understanding and semantic reasoning, the first phase integrates spatiotemporal context to infer the next app category and traffic level, and the second phase combines the user's long-term habits and short-term behavior to predict the specific app and traffic volume. This hierarchical strategy enables the LLM to model dependencies between the spatiotemporal trajectory and mobile app usage, thereby solving the \textbf{first} challenge.
Given that a long mobile app usage history is typically filled with redundant and weakly informative records, LLMApp introduces a prompt compressor to condense the long-term usage history into concise summary prompts that better capture the user's stable usage habits. By filtering out noise and retaining only the most salient behavioral cues, the prompt compressor reduces the computational burden for downstream LLM inference while preserving the key usage patterns required for practical reasoning, thereby alleviating the \textbf{second} challenge. 
Given the various modalities and structures of contextual information, LLMApp employs a context interpreter to encode diverse contextual signals into unified representations and project them into the LLM's semantic space. Therefore, the LLM can integrate spatiotemporal cues with app usage behaviors to uncover joint patterns and semantically correlated relationships within this shared semantic space, resolving the \textbf{third} challenge.

Overall, we summarize our contributions as follows:

$\bullet$ We investigate the use of large language models to predict mobile app usage. By enabling LLMs to reason about the interplay among location, time, and app category, our approach achieves more accurate and interpretable predictions of app usage, contributing to overall improvements in mobile service quality.

$\bullet$ We propose LLMApp, a hierarchical LLM-based model for predicting mobile app usage. LLMApp adopts a hierarchical predictor that captures the dependencies between spatiotemporal context and user app usage behavior, enabling the LLM to conduct coarse-to-fine semantic reasoning. To support this reasoning, LLMApp integrates a prompt compressor that condenses long, noisy usage histories into concise summary prompts. LLMApp employs a context interpreter to project diverse contextual signals into the LLM's embedding space.

$\bullet$ We conduct comprehensive experiments with two real-world datasets to validate our model's predictive performance. The results demonstrate that our model outperforms state-of-the-art baselines, achieving gains of over 9\% on critical metrics. Further in-depth analyses validate that our model successfully captures spatiotemporal patterns in app usage behavior and demonstrates strong scalability across diverse usage scenarios.

The remainder of this paper is organized as follows. Section~\ref{sec:preliminaries} presents preliminaries, including contextual impact on app usage, an overview of Large Language Models, and the problem definition. Section~\ref{sec:method} details the LLMApp framework, comprising its prompt compressor, context interpreter, and hierarchical predictor. Section~\ref{sec:exp} describes experimental setups and presents comprehensive results. Section~\ref{sec:related} discusses related work, and Section~\ref{sec:conclusion} concludes the paper.

%% file: Preliminaries.tex
\section{Preliminaries}
\label{sec:preliminaries}

\begin{figure}[tb]
\centering
\vspace{-2mm}
\includegraphics[width=0.5\textwidth]{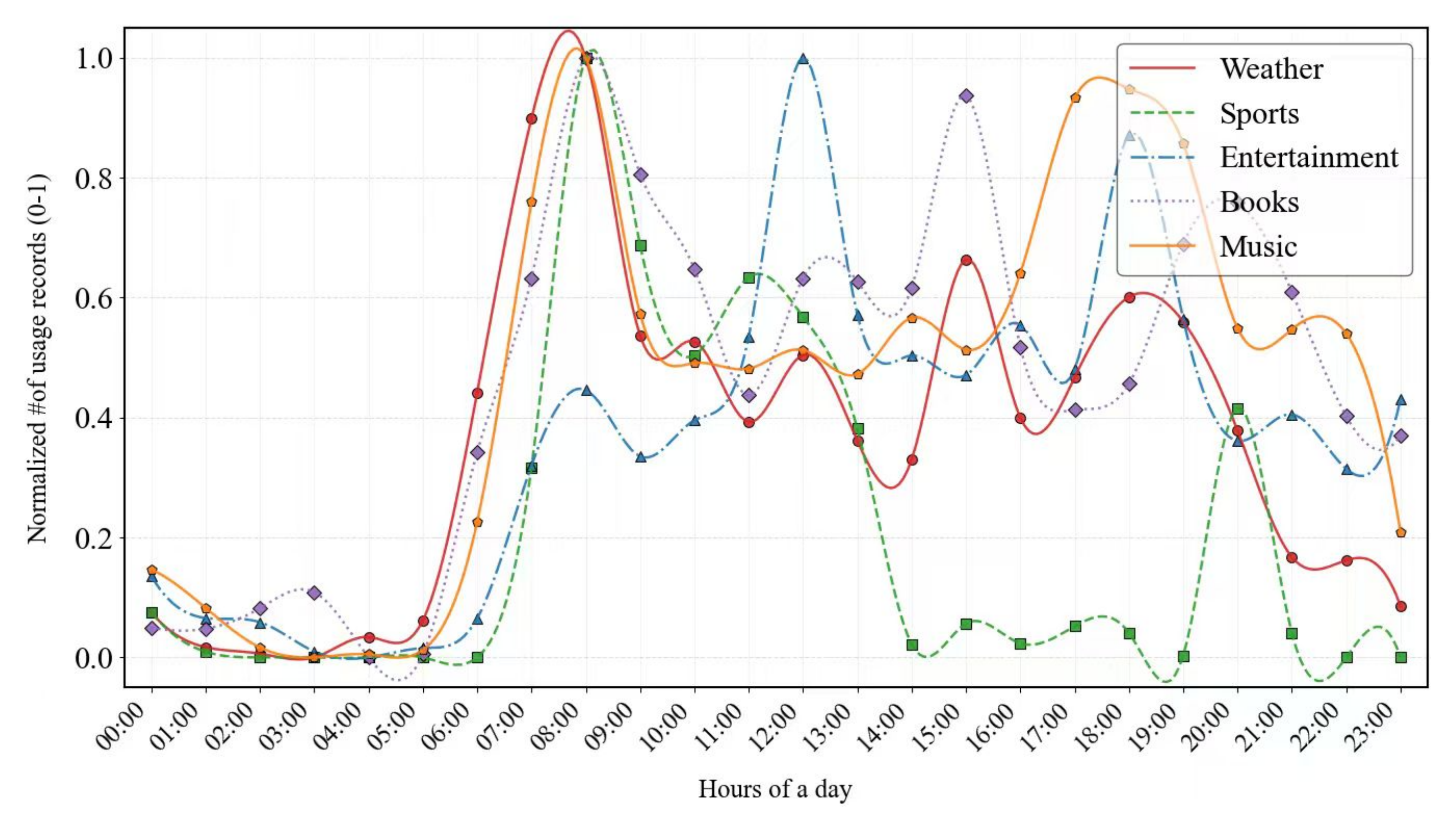}
\caption{App category usage records during a day based on shanghai dataset.}
\label{fig1:env}
\vspace{-2mm}
\end{figure}

\subsection{Contextual Impact on App Usage}

Prior studies~\cite{mehrotra2017understanding,signer2024exploring,jo2012spatiotemporal,li2021understanding,li2020extent,bohmer2011falling,xu2011identifying,wang2019modeling} consistently show that spatial and temporal context strongly shape app usage. Spatial context affects users' app preferences across different functional regions, while temporal context shapes their app choices based on daily routines. 
We also analyze a real-world dataset of mobile app usage collected in Shanghai.

\begin{figure}[tb]
\centering
\vspace{-2mm}
\includegraphics[width=0.5\textwidth]{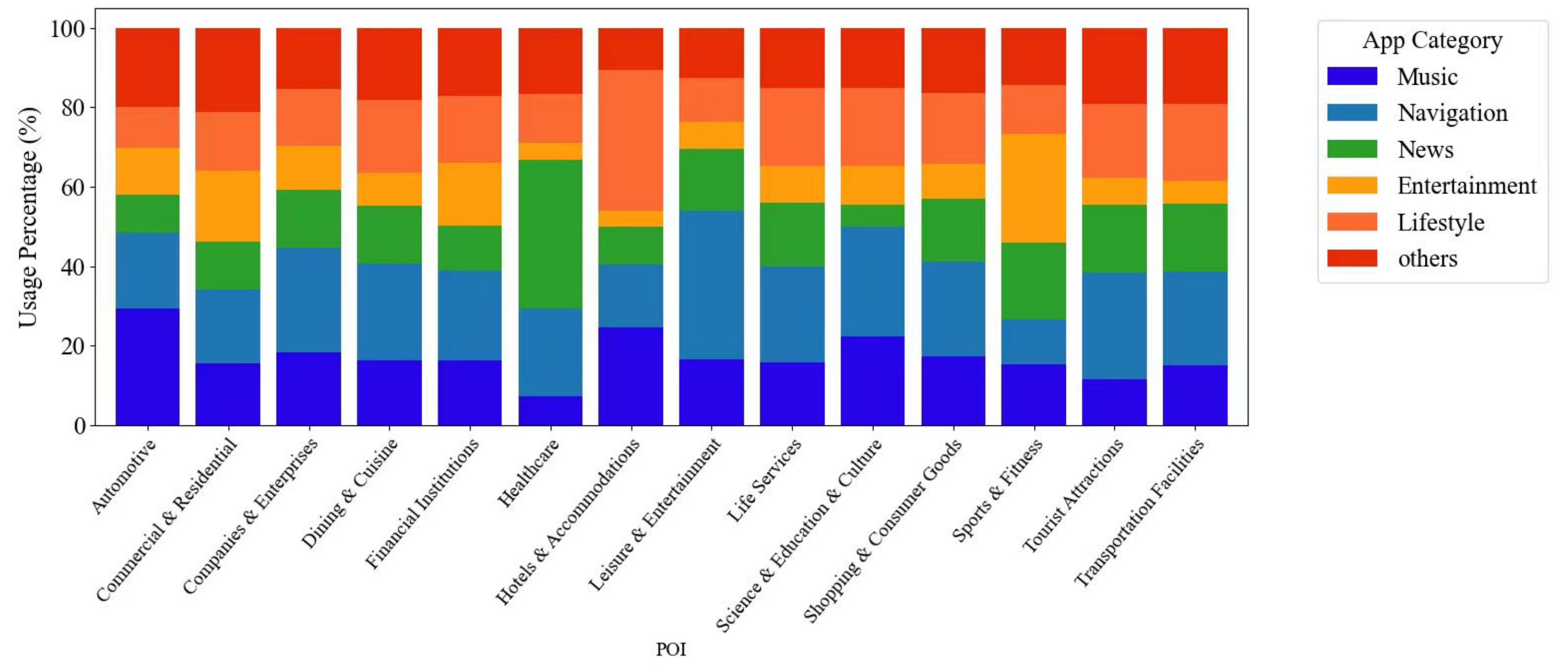}
\caption{App category usage records distribution in terms of locations based on shanghai dataset.}
\label{fig2:env}
\vspace{-2mm}
\end{figure}

We also analyze a real-world mobile app usage dataset collected in Shanghai (details of the dataset are introduced in Section~\ref{experiment_setup}) and observe phenomena similar to those described above. Fig.~\ref{fig1:env} shows the distribution of normalized usage records for five app categories according to the hour of the day. We can observe that users generally start accessing apps between 7 and 8 a.m. From there, the frequency steadily rises to multiple peaks at around 8 a.m., 11 a.m., 3 p.m., and 8 p.m., and then gradually decreases at night. Additionally, the utilization of different app categories varies significantly. Weather apps have a prominent peak around 8 a.m.; Sports apps show a peak around 8 a.m., and then a smaller one around 11 a.m.; Entertainment apps have a peak around 11 a.m. and another around 8 p.m.; Books apps have a more spread-out usage with minor peaks throughout the day, and Music apps have a peak around 8 a.m. and another around 8 p.m.
Furthermore, Fig.~\ref{fig2:env} illustrates the distribution of usage records for different app categories by location function. Navigation apps are more prevalent in transportation-related areas, while news apps are more frequently used in healthcare settings. Music apps also display distinct distribution proportions across various POIs. Therefore, we must incorporate spatio-temporal contextual information into the analysis of mobile users' app usage behavior to accurately reflect the complexity of their daily interactions with mobile devices.
These findings highlight the importance of modeling spatiotemporal context when predicting mobile app usage.

\subsection{Large Language Models}
Large Language Models (LLMs) are large-scale neural networks trained on massive text corpora to predict the next token in a sequence~\cite{wu2024continual,naveed2025comprehensive,plaat2024reasoning}. Through this training paradigm, LLMs acquire strong abilities in contextual understanding, semantic reasoning, and long-range dependency modeling. These capabilities allow LLMs to generalize beyond pure language modeling and to perform sequence reasoning across a wide range of inputs.
Beyond text generation, LLMs have been applied to structured prediction tasks such as next-event forecasting, user-intent modeling, and sequential decision making~\cite{makarov2025large,shi2023language,kim2025large,peng2025lc}. Recent studies further demonstrate that LLMs can be adapted to spatiotemporal domains, including trajectory prediction, mobility pattern inference, and human-activity reasoning~\cite{qin2025lingotrip,gong2024mobility}, highlighting their effectiveness in capturing dependencies, latent behavioral states, and cross-modal relationships.
Given these strengths, LLMs are well-positioned to model mobile app usage sequences by understanding user intent and spatiotemporal context. Initial attempts, such as MAPLE~\cite{khaokaew2024maple}, transform app usage records into natural-language prompts and use an LLM to infer the next app. These results indicate that LLMs have strong potential to advance mobile app usage prediction by leveraging their semantic reasoning and contextual abstraction capabilities.

\subsection{Problem Definition}
In this study, we aim to unleash the power of LLMs for predicting mobile app usage. Given historical app usage and contextual information, the goal of mobile app usage prediction is to estimate which app a user will open next and the expected traffic volume. 
Mathematically, an app usage record of a user $u$ can be represented as a tuple $s^u=(l^u, t^u, a^u, x^u)$, where $l^u$ denotes location, $t^u$ denotes timestamp, $a^u$ denotes the used app, and $x^u$ denotes the traffic volume. Given location $l^u_{T+1}$, time $t^u_{T+1}$, and a historical mobile app usage record sequence of length $T$:
\begin{equation}
\begin{split}
S^u &= \{s^u_1,s^u_2,\dots,s^u_{T}\} \\
    &= \{(l^u_1,t^u_1,a^u_1,x^u_1),(l^u_2,t^u_2,a^u_2,x^u_2),\dots,
       (l^u_{T},t^u_{T},a^u_{T},x^u_{T})\}.
\end{split}
\end{equation}
The goal of mobile app usage prediction is to design a model $G(\cdot)$ that predicts the next app and  traffic:
\begin{equation}
    (a^u_{T+1},x^u_{T+1})=G(S^u,l^u_{T+1},t^u_{T+1}).
\end{equation}

%% file: Methods.tex
\section{Methodology}
\label{sec:method}

\begin{figure*}[tb]
\vspace{-2mm}
\centering
\includegraphics[width=0.8\textwidth]{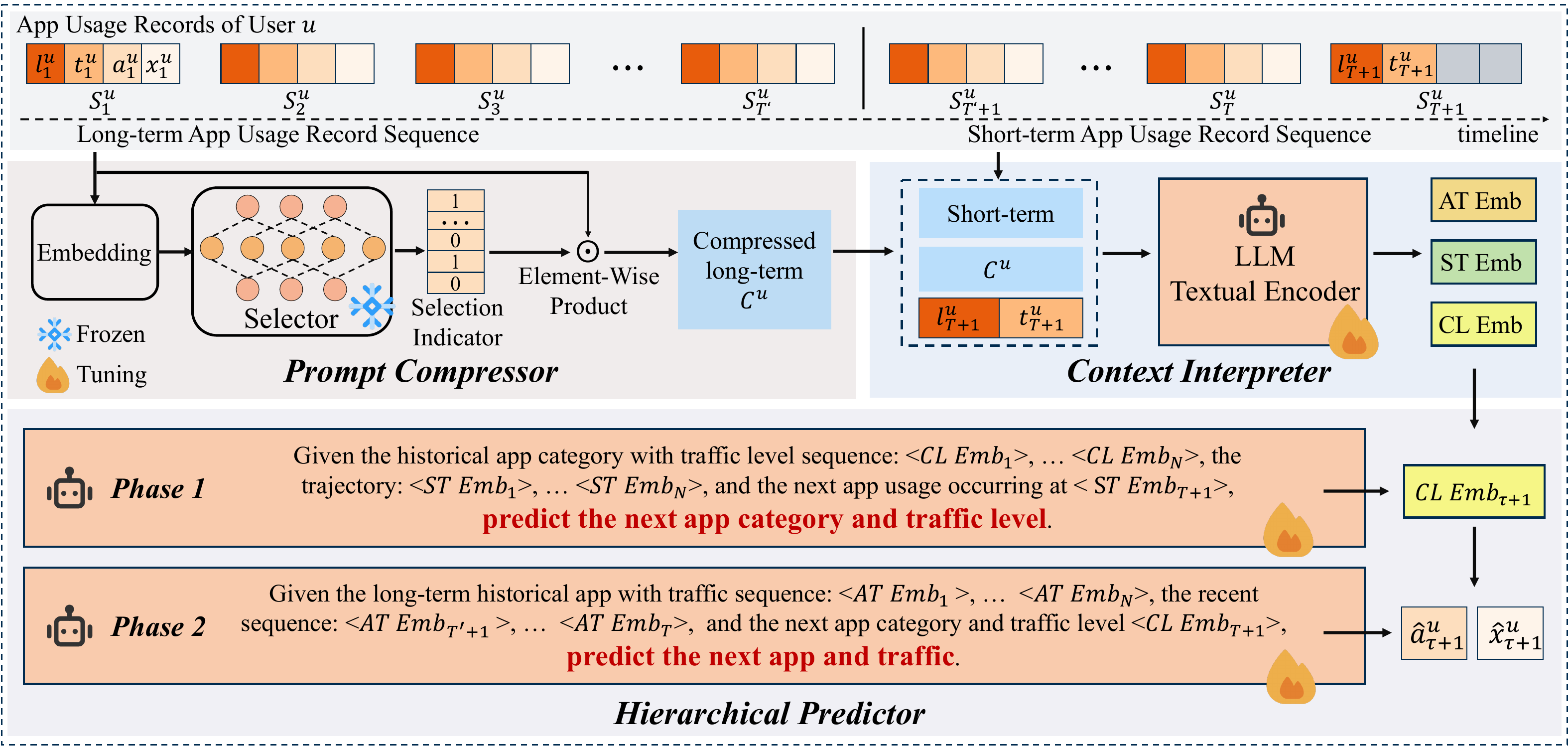}
\caption{Framework overview of LLMApp. LLMApp consists of three components: a prompt compressor for obtaining concise summary prompts, a context interpreter for encoding diverse contexts into representations within the LLM’s semantic space, and a hierarchical predictor for performing coarse-to-fine app usage prediction.}
\label{fig:framework}
\vspace{-2mm}
\end{figure*}

In this section, we present the design of LLMApp, describe its core components, and outline the mechanisms that enable effective prediction of mobile app usage.
\subsection{Framework Overview}
The overall architecture of LLMApp is illustrated in Fig.~\ref{fig:framework}, consisting of three key modules: a prompt compressor, a context interpreter, and a hierarchical predictor.
(1) \textbf{Prompt Compressor}. This module uses a self-supervised selector to filter redundant long-term mobile app usage logs, producing concise summary prompts that better reflect the user’s app usage habits. The compressor not only suppresses noise while preserving key usage patterns but also reduces the computational load for downstream LLM inference.
(2) \textbf{Context Interpreter}.This module maps diverse contextual information of app usage into unified representations and projects them into the LLM’s semantic space. The interpreter captures relationships between apps and traffic and maps spatiotemporal features and app usage behaviors into a shared embedding space, enabling effective modeling of joint patterns.
(3) \textbf{Hierarchical Predictor}.This module leverages an LLM-based hierarchical architecture to perform a coarse-to-fine mobile app usage prediction. The predictor first integrates spatiotemporal context to infer the next app category and traffic level. It further combines the user’s long-term habits and short-term behavior to predict the specific app and traffic volume.

\subsection{Prompt Compressor}
Long mobile app usage histories contain valuable signals for app prediction, such as stable user-specific usage habits. However, these logs are typically redundant and noisy, and directly converting them into prompts for LLM inference would introduce substantial distraction and unnecessary computational overhead. To address this, we introduce a prompt compressor that uses a self-supervised selector to extract a compact set of records that effectively represent user app usage habits. By suppressing noise and preserving salient behavioral patterns, the compressor generates a compressed long-term app usage record sequence that preserves the user’s salient usage patterns. The compressed sequence can then be transformed into a concise summary prompt that more accurately reflects the user’s app usage habits for downstream LLM inference.

To better capture both long-term user habits and recent behavioral signals of mobile app usage, we divide the historical app usage record sequence into long- and short-term parts. The short-term app usage record sequence refers to a recent, relatively short segment that reflects the current app usage state and the user’s immediate behavior. In contrast, the long-term app usage record sequence corresponds to a historically distant but much longer segment that preserves the user’s stable habits, such as preferences at specific locations and times and characteristic transition patterns between apps. Specifically, we use a threshold $T’$ to separate these two segments:
\begin{equation}
S^u_{long}={(s^u_1,s^u_2,\dots,s^u_{T’})},
\end{equation}
\begin{equation}
S^u_{short}={(s^u_{T’+1},s^u_{T’+2},\dots,s^u_{T})}.
\end{equation}

\begin{figure}[tb]
\vspace{-2mm}
\centering
\includegraphics[width=1.0\linewidth]{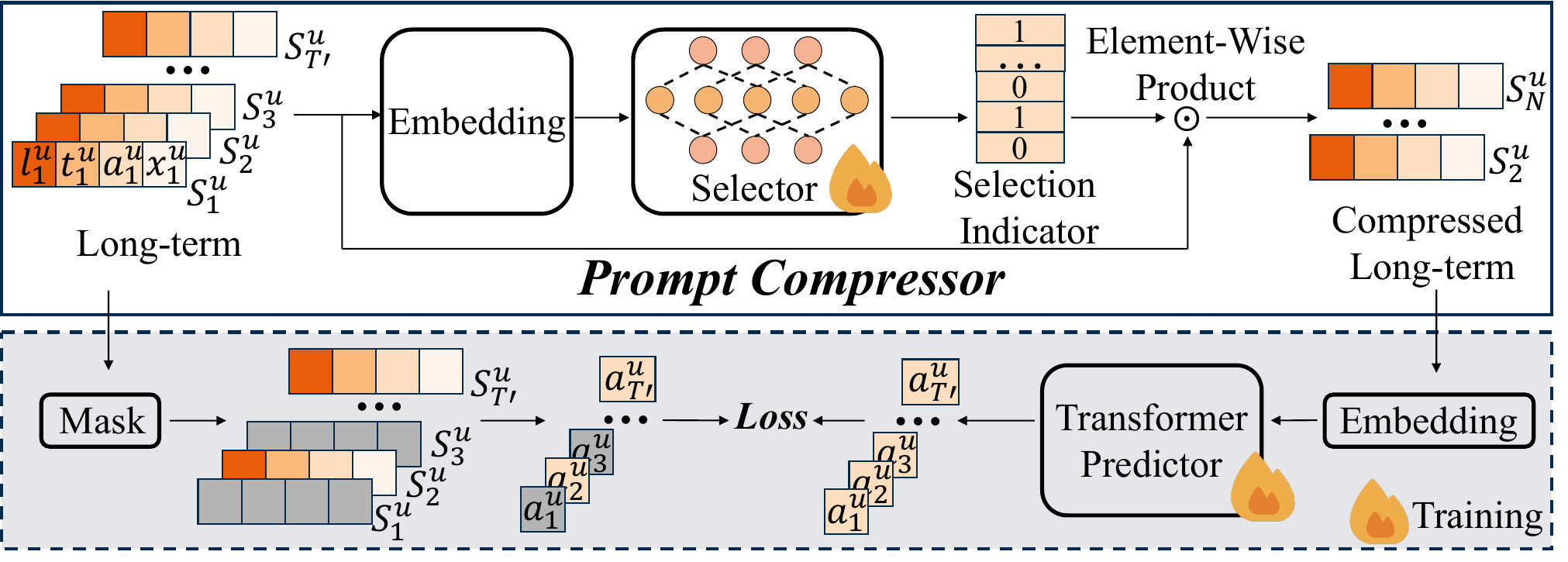}
\vspace{-2mm}
\caption{Self-supervised training of prompt compressor. The top shows how the prompt compressor embeds and condenses a long-term app usage record sequence to produce a compressed sequence for a summary prompt. The bottom illustrates self-supervised training, where a transformer predicts masked records to guide the selector in preserving informative usage patterns.}
\label{fig:compressor}
\vspace{-2mm}
\end{figure}

We propose a prompt compressor to select and compress a long-term sequence of app-usage records, thereby extracting meaningful user-habit signals. As shown in Fig.~\ref{fig:compressor}, the compressor first embeds all contextual features in the app usage. The feature embeddings are then concatenated and fed into a transformer architecture selector, which captures spatiotemporal dependencies in the long-term sequence. The selector assigns a relevance score to each position based on contextual correlations and selects $N$ positions with the highest scores. In addition, a selector indicator of the same length as $s^u_{long}$ is generated, with positions selected by the selector marked as one and all remaining positions set to zero. Finally, the compressed long-term app usage record sequence $C^u$ is obtained by element-wise multiplication of $S^u_{long}$ and the indicator.
Formally, let $M=(m_1,m_2,\dots,m_{T’}),~m_{\tau} \in (0,1)$ denote a selection of indices over the long-term app usage record sequence $S^u_{long}$, and the compressed long-term app usage record sequence is:
\begin{equation}
C^u=S^u_{long} \odot M.
\end{equation}
For clarity, this corresponds to keeping only the selected elements in $S^u_{long}$. Specifically,
\begin{equation}
C^u={c^u_{1},\dots,c^u_{N}},\quad c^u_{n}=s^u_{\tau_i}, \quad \tau_i \in {\tau|m_{\tau}=1}.
\end{equation}

The selector is trained in a self-supervised manner, and the core insight is that the optimal selection of app usage records should maximize predictive information about user context.
\textcolor{black}{Formally, $Q$ denotes the target user context.} The optimal selector should maximize the mutual information between the compressed long-term app usage record sequence $C^u$ and $Q$:
\begin{equation}
\underset{p_{\psi}}{max}I(C^u;Q),\quad \text{ s.t. } M\sim p_{\psi }(M\mid S^u_{long}),
\end{equation}
where $p_{\psi}$ denotes the selector with parameter $\psi$, $I$ quantifies the contextual relevance through mutual information maximization. Using the standard expansion $I(C^u;Q)=H(Q)-H(Q \mid C^u)$, where $H(\cdot)$ is the entropy, this objective is equivalent, up to a constant, to maximizing the conditional log-likelihood $\mathbb{E}[\log p(Q \mid C^u)]$. Since the true conditional distribution is intractable, we introduce a transformer predictor $q_{\theta}(Q \mid C^u)$ with parameter $\theta$ and optimize the following variational lower bound:
\begin{equation}
\underset{\psi,\theta}{max}\mathbb{E}{S^u{long},Q}\mathbb{E}{M \sim p{psi}(M \mid S^u_{long})}[\log q_{\theta}(Q \mid C^u)].
\end{equation}
In training the long-term data compressor, Q denotes the set of masked app IDs, and $q_{\theta}$ factorizes over masked positions. This leads to a BERT-style masked reconstruction loss, computed only on masked tokens, with the predictor attending primarily to the key patterns selected by the selector.
\textcolor{black}{Through this self-supervised selection process, the prompt compressor learns to identify records that contain strong predictive information about the user’s app usage behavior. Since these selected records are optimized to reconstruct masked usage signals from long-term histories, they tend to preserve stable preferences, recurrent transition patterns, and representative spatiotemporal usage contexts. Therefore, the compressed sequence can serve as a compact yet informative summary of user habits for downstream LLM reasoning.}

\begin{table*}[t]
\centering
\small
\caption{Examples of contextual prompt construction, generated textual descriptions, and corresponding unified embeddings.}
\begin{tabular}{>{\centering\arraybackslash}m{2.9cm}
                |>{\centering\arraybackslash}m{5.5cm}
                |>{\centering\arraybackslash}m{6.5cm}
                |>{\centering\arraybackslash}m{1.5cm}}
\hline
\textbf{Input} & \textbf{Prompt} & \textbf{Generated Textual Description} & \textbf{Embedding} \\
\midrule
\makecell{$l^u_{\tau}=1283$ \\ $t^u_{\tau}=\text{May 14, 18:37}$}
&
Describe the spatiotemporal context: \textit{Tuesday 18:37}, POI distribution nearby  includes: \textit{Transportation 32\%, Shopping 18\%, Dining 15\%,\dots}.
&
The user is in a transportation-dominant area with surrounding commercial and dining activities. The time is early evening on a weekday, typically associated with commuting or post-work usage routines.
&
$\langle Emb^{ST}_{u,\tau} \rangle$
\\
\midrule
\makecell{$Cate(a^u_{\tau})=\text{Travel}$\\$Lev(x^u_{\tau})=\text{medium}$}
&
Describe semantic meaning of a mobile app usage: \textit{app category: Travel, traffic level: medium}.
&
This usage corresponds to a travel-related application with a medium data demand, commonly linked to navigation or mobility-related queries.
&
$\langle Emb^{CL}_{u,\tau} \rangle$
\\
\midrule
\makecell{$a^u_{\tau}=207$\\$x^u_{\tau}=7.7420\text{ MB}$}
&
Interpret the following app usage event: \textit{app \#207, traffic: 7.7420 MB}.
&
The record indicates that app was accessed with a moderate traffic volume, suggesting a lightweight request such as fetching updates or loading a small set of resources.
&
$\langle Emb^{AT}_{u,\tau} \rangle$
\\
\hline
\end{tabular}
\label{tab:prompt}
\end{table*}

\subsection{Context Interpreter}
Unleashing the power of LLMs for mobile app usage requires the LLM to understand how diverse contextual signals, such as location, time, app category, and traffic level, jointly influence user behavior. However, these signals differ greatly in modality and structure, making them incompatible with the linguistic representation space in which LLMs operate. To bridge this gap, LLMApp introduces a context interpreter that encodes diverse contextual inputs into unified representations and projects them into the LLM’s semantic space. This unified embedding space enables the LLM to integrate spatiotemporal cues with app usage behaviors and perform coherent joint reasoning over their relationships.

We integrate contextual information as follows. (1) Spatiotemporal trajectories have contextual impacts on app usage, as users exhibit different usage patterns across locations and times. We therefore organize spatiotemporal features by representing the functional characteristics of the current location through the distribution of nearby POI types and by encoding temporal context. (2) The category of the used app and its traffic level show latent correlations with these spatiotemporal contextual features, and incorporating category and level predictions as intermediate conditions further improves the prediction performance for the next app and its traffic. We assign each app to a category (e.g., travel, utilities, music) and derive traffic levels by computing the empirical traffic distribution and partitioning it into ten bins. (3) The specific app ID and traffic volume are jointly encoded to represent the app usage behavior.

We construct three types of prompts and use an LLM as a textual encoder to embed the corresponding contextual features. As illustrated in Table~\ref{tab:prompt}, we build prompts using location and time information and feed them into the context interpreter to obtain a joint embedding in the unified semantic space, denoted as the Spatial-Temporal Embedding ($Emb^{ST}{u,\tau}$). Similarly, we generate the Category-Level Embedding ($Emb^{CL}{u,\tau}$) from app category and traffic level, and the App-Traffic Embedding ($Emb^{AT}{u,\tau}$) from the specific app ID and traffic volume. Formally, the embeddings are defined as:
\begin{align}
Emb^{ST}{u,\tau} &= Encoder!\left(\rho!\left(POI(l^u_{\tau}),, t^u_{\tau}\right)\right), \
Emb^{CL}{u,\tau} &= Encoder!\left(\rho!\left(Cate(a^u{\tau}),, Lev(x^u_{\tau})\right)\right), \
Emb^{AT}{u,\tau} &= Encoder!\left(\rho!\left(a^u{\tau},, x^u_{\tau}\right)\right).
\end{align}
where $POI(\cdot)$ denotes the POI profile of the given location, $\rho(\cdot)$ denotes the prompt construction function, $Cate(\cdot)$ returns the category of an app, and $Lev(\cdot)$ maps the traffic volume to the corresponding traffic-level bin. Noticeably, the encoded embeddings of the compressed long-term app usage record sequence $C^u={c^u_1,\dots,c^u_N}$ are indexed accordingly as $Emb^{ST}{u,n}$, $Emb^{CL}{u,n}$, and $Emb^{AT}_{u,n}$, where $k=1,\dots, N$.

\subsection{Hierarchical Predictor}
Predicting mobile app usage requires capturing how contextual factors jointly influence users’ app usage behavior. However, enabling an LLM to learn these dependencies directly is challenging, as complex interactions across locations, times, and user-specific habits shape such behavioral patterns.
To solve this challenge, we propose a hierarchical predictor that decomposes the prediction process into two phases and leverages an LLM's strong semantic understanding and reasoning capabilities to capture higher-order interactions in contextual signals and predict mobile app usage.
Specifically, in the first phase, the predictor focuses on inferring the next app category and traffic level by modeling the dependencies between spatiotemporal context and mobile app usage. In the second phase, based on the inferred next app category and traffic level, the predictor integrates the user’s long-term habits and recent app usage behavior to produce a final prediction of the specific app and traffic volume. This hierarchical structure guides the LLM to learn the underlying associations between spatiotemporal trajectories and app usage behavior.

At the first stage, the predictor learns the dependencies between spatial context and mobile app usage, integrating spatiotemporal contextual signals to estimate the next app category and traffic level. The app usage exhibits distinct patterns in both category choice and traffic volume across locations and times. Learning the dependencies between the spatiotemporal trajectory and the app usage category enables the model to predict the next app category and traffic level as an intermediate condition for the final prediction of the specific app and traffic volume. As a result, this intermediate modeling step improves the overall prediction accuracy. The predictor reasons about how location and time interact with the user’s choice of app category and traffic level, and predicts the next app category and traffic level.
Specifically, the predictor receives three types of unified embeddings: (1) $Emb^{ST}{u,n}$, representing the selected spatiotemporal trajectory of the user, (2) $Emb^{CL}{u,n}$, derived from historical app category and traffic level records, and (3) $Emb^{ST}{u, T+1}$, the spatiotemporal context embedding of the future usage position.
As illustrated in Fig.~\ref{fig:framework}, we organize these embeddings into a prefix-style contextual prompt and inject them into the LLM as conditioning. Guided by this context-aware prefix, the LLM captures the dependencies between the user’s spatiotemporal trajectory and category-level usage behavior. It produces an embedding of the following app category and its traffic level, denoted as $Emb^{CL}{u, T+1}$.

In the second stage, the predictor first extracts sequential dependencies from the compressed long-term app usage record, thereby revealing the user’s app usage habits. Then, the predictor further integrates the predicted next app category and traffic level, along with recent behavioral signals, to infer the specific next app and its traffic volume, enabling accurate mobile app usage prediction.
The compressed long-term sequence $C^u$ provides a concise summary of the user’s behavioral patterns, including preferred apps, app transition tendencies, and typical traffic usage. In contrast, the short-term sequence reflects the user’s recent behavior and current usage context, capturing situational demands for app usage. Combining long-term usage habits with short-term behavior enables the predictor to better understand the user’s app usage pattern, thereby improving the accuracy of the final prediction. At this stage, the predictor receives three embedding signals: (1) $Emb^{CL}{u, T+1}$, the inferred next app category and traffic level from the first stage, (2) $Emb^{AT}{u,n}$, the long-term app-usage habits extracted from the compressed historical sequence, and (3) $Emb^{AT}_{u,\tau}$, where $\tau>T’$, the short-term app-usage behavior that captures the user’s most recent actions. Similarly, these signals are assembled into a prefix-style contextual prompt and fed into the LLM, generating the final prediction of the next app and its corresponding traffic volume.

\begin{table*}[tb]
\centering
\caption{Dataset Statistics.}
\footnotesize
\resizebox{1.0\linewidth}{!}{
\begin{tabular}{c|ccccccc}
\toprule
Cities & Collection Date & \# Users & \# Apps & \# Records & \makecell{\# App \\ Categories} & \# Location & \makecell{Avg. Trajectory \\Length per day}\\
\midrule
Shanghai & \text{19 April 2016 -- 26 April 2016} & 870  & 305 & $2,518,108$ & 20 & $12,576$ & 16.3 \\
Nanchang & \text{19 May 2022 -- 25 May 2022}     & 9466 & 640 & $20,556,868$ & 22 & $63,200$ & 17.1 \\
\bottomrule
\end{tabular}
}
\label{tab:Dataset}
\end{table*}

We apply a prefix-tuning strategy to fine-tune the LLM. In this design, conditional embeddings are projected into soft prefix vectors that are inserted at the beginning of the LLM’s hidden sequence, steering the model’s attention toward signals and enabling it to internalize contextual features, personalized usage patterns, and recent app usage behaviors during prediction. During training, the encoder and predictor are jointly optimized end-to-end, while the selector is kept frozen. The objective combines losses for the next-app category, app ID, and traffic volume. The app ID prediction is supervised with a CE (cross-entropy) loss $\mathcal{L}{app}$ that measures the difference between the predicted app probability distribution $logits{app}$ and the ground truth $a^u_{T+1}$:
\begin{equation}
\mathcal{L}{app}=CE(logits{app},a^u_{T+1}).
\end{equation}
Because uncertainty in the next app category inference can lead to significant drift in the app ID prediction, we design the predictor to produce multiple next app category candidates and optimize them with CE loss $ \ mathcal {L} _ {cate} $ to improve the reliability of the final app ID prediction:
\begin{equation}
\mathcal{L}{cate}=CE(logits_{cate},Cate(a^u_{T+1})),
\end{equation}
where $logits_{cate}$ denotes the predicted distribution of subsequent app categories; for traffic prediction, the predictor outputs a non-negative traffic value via a softplus transformation. The heavy-tailed nature of traffic volume is handled by clipping the ground-truth traffic at the 95th percentile, and a smooth L1 loss is applied in log-space to stabilize training on long-range values, while an additional linear-space error term keeps the model responsive to larger traffic magnitudes:
\begin{equation}
\mathcal{L}_{\text{traf-log}}= \mathrm{Huber}\!\left(\log(1+\hat{x}^u_{T+1}),\, \log(1+x^u_{T+1})\right),
\end{equation}
\begin{equation}
    \mathcal{L}_{\text{traf-lin}}=\lambda\, \big|\hat{x}^u_{T+1} - x^u_{T+1}\big|,
\end{equation}
\begin{equation}
    \mathcal{L}_{\text{traf}}=\mathcal{L}_{\text{traf-log}}+\lambda_{\text{traf-lin}}\mathcal{L}_{\text{traf-lin}},
\end{equation}
\textcolor{black}{where the $\lambda_{\text{traf-lin}}$ is an adjustable parameter.} The overall training objective is:
\begin{equation}
\mathcal{L}=\mathcal{L}_{app}+\lambda_{cate}\mathcal{L}_{cate}+\lambda_{traf}\mathcal{L}_{traf},
\end{equation}
where $\lambda_{cate}$ and $\lambda_{traf}$ are adjustable parameters.

%% file: Experiments.tex
\section{Experiments}
\label{sec:exp}

In this section, we evaluate LLMApp's performance on the app usage prediction task through comprehensive experiments. We first describe the datasets and experimental settings. Then we present results from a series of experiments, including an overall performance comparison, ablation studies, app category prediction analysis, data compression case studies, LLM backbone comparisons, hyperparameter analysis, generalization evaluation, spatial dependency modeling assessment, and traffic prediction performance.

\subsection{Experimental Setup}
\label{experiment_setup}

\subsubsection{Datasets}
We evaluate the performance of LLMApp on the Shanghai and Nanchang mobile app usage datasets.
The Shanghai dataset was collected by an internet service provider over one week in April 2016 and includes the entire metropolitan region of Shanghai, one of the largest cities in the world. 
The Nanchang dataset was collected by one of the largest network operators in China between May 19 and 25, 2022, covering the entire metropolitan area of Nanchang, a capital city in China. Both datasets contain fine-grained records of mobile app usage, with each entry including the corresponding location, timestamp, app ID, and traffic volume. Table~\ref{tab:Dataset} summarizes the statistics of the two real-world datasets. In this paper, we randomly split each dataset into 80\% for training, 10\% for validation, and 10\% for testing, ensuring no overlap across splits.
\textcolor{black}{
All datasets used in this study were collected with proper authorization from users and data providers, and were anonymized and desensitized before research use. Specifically, user identities were replaced with anonymized IDs, raw geographic coordinates were transformed into coarse-grained region IDs, and exact usage timestamps were discretized into hourly intervals. In addition, the datasets were used strictly for academic research purposes, and were not used for commercial profiling, individual identification, advertising, or any decision-making process targeting specific users. Therefore, researchers cannot access private user identities, precise locations, or personally identifiable information, which helps reduce privacy risks and prevent potential misuse of predictive user profiles.
}

\subsubsection{Settings}
In our experiments, we use LLaMA-2-7B as the backbone of LLMApp and fine-tune through LoRA. The selector employs 128-dimensional embeddings, a 256-dimensional hidden layer, and a key-pattern length of 12. The predictor uses a long-term window of 128 records and a short-term window of 12 records.
\textcolor{black}{In terms of traffic-level construction, we group app usage records by app category and collect the corresponding traffic volumes from the training set. To reduce the influence of long-tailed distributions and extreme values, we apply a logarithmic transformation to the traffic volumes. Then, for each app category, the transformed traffic distribution is divided into ten intervals, corresponding to ten traffic levels.}
Model optimization follows AdamW with a learning rate of $1\times10^{-4}$ and zero weight decay, using batch sizes of 128 for the selector and 32 for the predictor. All training and inference are carried out on a single NVIDIA A100 40GB PCIe GPU.

\begin{table*}[tb]
\centering
\footnotesize
\caption{App ID Prediction results compared with baselines on two real-world \textcolor{black}{datasets}. Bold indicates the best overall result; underline denotes the second-best methods.}
\label{tab:baseline}
\resizebox{0.85\linewidth}{!}{
\begin{tabular}{c|ccccccccc}
\toprule
\multicolumn{1}{c|}{\textbf{Cities}} & \multicolumn{9}{c}{\textbf{Shanghai}} \\
\midrule
\multirow{2}{*}{\raisebox{-0.3\height}{\textbf{Methods}}}
  & \multicolumn{3}{c}{\textbf{Acc@K}} 
  & \multicolumn{3}{c}{\textbf{MRR@K}} 
  & \multicolumn{3}{c}{\textbf{NDCG@K}} \\
  \cmidrule(lr){2-4}\cmidrule(lr){5-7}\cmidrule(lr){8-10}
& 1 & 5 & 10 & 1 & 5 & 10 & 1 & 5 & 10 \\
\midrule
APPUsage2Vec & 0.2555 & 0.5676 & 0.6976 & 0.2555 & 0.3668 & 0.3844 & 0.2555 & 0.4167 & 0.4589 \\
CAP & 0.1760 & 0.3006 & 0.3599 & 0.1760 & 0.2208 & 0.2282 & 0.1760 & 0.2407 & 0.2582 \\
SA-GCN & 0.2422 & 0.4470 & 0.5189 & 0.2422 & 0.3208 & 0.3305 & 0.2422 & 0.3524 & 0.3755 \\
LSTM & 0.2056 & 0.4183 & 0.4982 & 0.2056 & 0.2816 & 0.2927 & 0.2056 & 0.3156 & 0.3413 \\
DUGN & 0.2641 & 0.5852 & 0.7208 & 0.2641 & 0.3778 & 0.3960 & 0.2641 & 0.4292 & 0.4727 \\
SHARE & 0.1996 & 0.4329 & 0.5371 & 0.1996 & 0.2970 & 0.3009 & 0.1996 & 0.3233 & 0.3571 \\
CADHG & 0.3980 & \underline{0.9191} & \underline{0.9478} & 0.3980 & 0.6104 & 0.6141 & 0.3980 & 0.6889 & 0.6977 \\
MAPLE & \underline{0.5850} & 0.9183 & 0.9395 & \underline{0.5850} & \underline{0.6962} & \underline{0.7015} & \underline{0.5850} & \underline{0.7678} & \underline{0.8380} \\
\textbf{LLMApp} & \textbf{0.6382} & \textbf{0.9306} & \textbf{0.9480} &
\textbf{0.6382} & \textbf{0.7084} & \textbf{0.7153} &
\textbf{0.6382} & \textbf{0.8017} & \textbf{0.8485} \\
\midrule
\textit{Improvement} &
\textbf{+9.09\%} &
\textbf{+1.34\%} &
\textbf{+0.90\%} &
\textbf{+9.09\%} &
\textbf{+1.75\%} &
\textbf{+1.97\%} &
\textbf{+9.09\%} &
\textbf{+4.42\%} &
\textbf{+1.25\%} \\
\midrule
\multicolumn{1}{c|}{\textbf{Cities}} & \multicolumn{9}{c}{\textbf{Nanchang}} \\
\midrule
\multirow{2}{*}{\raisebox{-0.3\height}{\textbf{Methods}}}
  & \multicolumn{3}{c}{\textbf{Acc@K}} 
  & \multicolumn{3}{c}{\textbf{MRR@K}} 
  & \multicolumn{3}{c}{\textbf{NDCG@K}} \\
  \cmidrule(lr){2-4}\cmidrule(lr){5-7}\cmidrule(lr){8-10}
& 1 & 5 & 10 & 1 & 5 & 10 & 1 & 5 & 10 \\
\midrule
APPUsage2Vec & 0.2320 & 0.7402 & 0.8972 & 0.2320 & 0.4050 & 0.4268 & 0.2320 & 0.4879 & 0.5393 \\
CAP & 0.1877 & 0.4275 & 0.5887 & 0.1877 & 0.2687 & 0.2897 & 0.1877 & 0.3078 & 0.3594 \\
SA-GCN & 0.2870 & 0.6815 & 0.7402 & 0.2870 & 0.4347 & 0.4428 & 0.2870 & 0.4964 & 0.5156 \\
LSTM & 0.2397 & 0.7306 & 0.8517 & 0.2397 & 0.4162 & 0.4316 & 0.2397 & 0.4898 & 0.5274 \\
DUGN & 0.2390 & 0.7624 & 0.8992 & 0.2390 & 0.4171 & 0.4396 & 0.2390 & 0.5026 & 0.5550 \\
SHARE & 0.2543 & 0.7336 & 0.8596 & 0.2543 & 0.4371 & 0.4443 & 0.2543 & 0.5043 & 0.5443 \\
CADHG & 0.3894 & 0.8678 & \underline{0.9266} & 0.3894 & \underline{0.5689} & \underline{0.5825} & 0.3894 & \underline{0.6031} & \underline{0.6463} \\
MAPLE & \underline{0.4729} & \underline{0.8732} & 0.9245 & \underline{0.4729} & 0.5854 & 0.5875 & \underline{0.4729} & 0.5971 & 0.6402 \\
\textbf{LLMApp} &
\textbf{0.5040} & \textbf{0.8810} & \textbf{0.9350} &
\textbf{0.5040} & \textbf{0.5923} & \textbf{0.5936} &
\textbf{0.5040} & \textbf{0.6446} & \textbf{0.6819} \\
\midrule
\textit{Improvement} &
\textbf{+6.58\%} &
\textbf{+0.89\%} &
\textbf{+1.14\%} &
\textbf{+6.58\%} &
\textbf{+1.18\%} &
\textbf{+1.04\%} &
\textbf{+6.58\%} &
\textbf{+6.88\%} &
\textbf{+5.51\%} \\
\bottomrule
\end{tabular}
}
\vspace{-3mm}
\end{table*}

\subsubsection{Metrics}
We compare the predicted top-K list against the ground truth to evaluate the app prediction performance of our model, and we employ three commonly used evaluation metrics: $Acc@K$, $MRR@K$, and $NDCG@K$. For traffic prediction accuracy assessment, we report four regression metrics: Mean Absolute Error (MAE), Mean Squared Error (MSE), Root Mean Squared Error (RMSE), and Spearman's Rank Correlation Coefficient (Spearman).

\textbf{Accuracy($\mathbf{Acc@K}$)} measures the proportion of test samples whose ground-truth app appears in the top-K predicted list. It is defined as:
\begin{equation}
    Acc@K = \frac{\# hit@K}{|N_{test}|},
\end{equation}
where $\# hit@K$ represents the number of hits in the test dataset and $|N_{test}|$ signifies total quantity of the test dataset.

\textbf{Mean Reciprocal Rank ($\mathbf{MRR@K}$)} gives better points to hits that are located at the top of the list. It is computed as:
\begin{equation}
    MRR@K = \frac{1}{N_{test}} \sum_{i=1}^{N_{test}}rank_i,
\end{equation}
where $rank_i$ is the reciprocal of the rank for the $i$th prediction.

\textbf{Normalized Discounted Cumulative Gain ($\mathbf{NDCG@K}$)} is to add the gain of all results starting at the top of the ranking list where the results at lower ranks are discounted, and it can be calculated as: 
\begin{equation}
NDCG@K=
\frac{
\sum_{i=1}^{K} \frac{2^{rel_i}-1}{\log_2(i+1)}
}{
\sum_{j=1}^{|REL_K|} \frac{rel_j}{\log_2(j+1)}
}.
\end{equation}
where $rel_i$ denotes the graded relevance of the result at position $i$, and $REL_K$ means the list of predictions in the result ranking list up to position K. 

\textbf{Spearman's Rank Correlation Coefficient (Spearman)} measures the monotonic relationship between two continuous variables by comparing the rank order of their values. It is defined as:
\begin{equation}
\rho = \frac{cov(R(X),R(Y))}{\sigma_{R(X)}\sigma_{R(Y)}},
\end{equation}
where $X$ and $Y$ denote the predicted and ground-truth traffic values, and $R(X)$ and $R(Y)$ are their ranks. For $Acc@K$,$MRR@K$,$NDCG@K$, and Spearman, larger values indicate better app prediction performance. For MAE, MSE, and RMSE, smaller values are preferred.

\subsubsection{Baselines}

We evaluate the predictive performance of LLMApp in comparison with baseline methods outlined below:

\textbf{AppUsage2Vec~\cite{zhao2019appusage2vec}} computes each app’s contribution to the target app through an attention mechanism, then applies a dual-DNN module to learn user-specific app usage characteristics, and finally predicts the next app based on the inferred personalized usage patterns.

\textbf{CAP~\cite{chen2019cap}} builds a graph from apps and contextual signals, embeds their relationships, derives user preference profiles, and then performs app usage prediction through a graph constructed from all users’ historical records. 

\textbf{SA-GCN~\cite{yu2020semantic}} converts app usage with side information into a graph and applies a graph convolution network to learn semantic-aware representations for downstream app usage prediction.

\textbf{LSTM~\cite{hochreiter1997long}} models temporal dependencies in user behavior through recurrent connections, capturing current user intent and using it to predict the next app.
 
\textbf{DUGN~\cite{ouyang2022learning}} constructs a dynamic app usage graph from prior behavior, extracts user intent with graph neural networks, and fuses dynamic intent with current needs to forecast the next app.

\textbf{SHARE~\cite{wang2021session}} builds hypergraphs to encode contextual signals within sessions, applies a hypergraph attention network to infer user intent, and integrates general and current intent via self-attention for next-item prediction.

\begin{table*}[tb]
\centering
\footnotesize
\caption{Evaluation results of different variants of LLMApp on two real-world datasets.}
\label{tab:ablation}
\resizebox{0.85\linewidth}{!}{
\begin{tabular}{p{2.3cm}|ccccccccc}
\toprule
\multicolumn{1}{c|}{\textbf{Cities}} & \multicolumn{9}{c}{\textbf{Shanghai}} \\
\midrule
\multirow{2}{*}{\raisebox{-0.3\height}{\textbf{Ablation Variant}}}
  & \multicolumn{3}{c}{\textbf{Acc@K}} 
  & \multicolumn{3}{c}{\textbf{MRR@K}} 
  & \multicolumn{3}{c}{\textbf{NDCG@K}} \\
  \cmidrule(lr){2-4}\cmidrule(lr){5-7}\cmidrule(lr){8-10}
& 1 & 5 & 10 & 1 & 5 & 10 & 1 & 5 & 10 \\
\midrule
w/o hierarchical & 0.4583 & 0.7103 & 0.7900 & 0.4583 & 0.5492 & 0.5600 & 0.4583 & 0.5892 & 0.6152 \\
w/o app category & 0.4830 & 0.7307 & 0.8042 & 0.4830 & 0.5781 & 0.5888 & 0.4830 & 0.6163 & 0.6409 \\
w/o user habits & 0.5676 & 0.8177 & 0.9104 & 0.4676 & 0.6609 & 0.6734 & 0.5676 & 0.7000 & 0.7302 \\
LLMApp & \textbf{0.6382} & \textbf{0.9306} & \textbf{0.9480} &
\textbf{0.6382} & \textbf{0.7084} & \textbf{0.7153} &
\textbf{0.6382} & \textbf{0.8017} & \textbf{0.8485} \\
\midrule

\multicolumn{1}{c|}{\textbf{Cities}} & \multicolumn{9}{c}{\textbf{Nanchang}} \\
\midrule
\multirow{2}{*}{\raisebox{-0.3\height}{\textbf{Ablation Variant}}}
  & \multicolumn{3}{c}{\textbf{Acc@K}} 
  & \multicolumn{3}{c}{\textbf{MRR@K}} 
  & \multicolumn{3}{c}{\textbf{NDCG@K}} \\
  \cmidrule(lr){2-4}\cmidrule(lr){5-7}\cmidrule(lr){8-10}
& 1 & 5 & 10 & 1 & 5 & 10 & 1 & 5 & 10 \\
\midrule
w/o hierarchical & 0.4072 & 0.6303 & 0.7371 & 0.4072 & 0.4666 & 0.4802 & 0.4072 & 0.5078 & 0.5418 \\
w/o app category & 0.4424 & 0.7079 & 0.8167 & 0.4424 & 0.5451 & 0.5592 & 0.4424 & 0.5859 & 0.6208 \\
w/o user habits & 0.4238 & 0.6713 & 0.8069 & 0.4238 & 0.5026 & 0.5230 & 0.4238 & 0.5446 & 0.5881 \\
LLMApp &
\textbf{0.5040} & \textbf{0.8810} & \textbf{0.9350} &
\textbf{0.5040} & \textbf{0.5923} & \textbf{0.5936} &
\textbf{0.5040} & \textbf{0.6446} & \textbf{0.6819} \\
\bottomrule
\end{tabular}
}
\end{table*}

\textbf{CADHG~\cite{huang2025predicting}} builds context-aware dynamic hypergraphs from users’ spatiotemporal trajectories and app sessions, applies graph neural networks to capture evolving user intents, and predicts the next app by fusing historical usage graphs with current context.

\textbf{MAPLE~\cite{khaokaew2024maple}} exploits large language models to embed diverse contextual data and installed-app similarity, addresses the cold-start problem, and predicts mobile app usage by mapping all available context and usage information into a unified semantic space.

\subsection{Experimental Results}
\subsubsection{Overall Performance Evaluation}
To evaluate the effectiveness of our proposed LLMApp model, we benchmark it against leading state-of-the-art (SOTA) baselines on two real-world datasets. Table~\ref{tab:baseline} summarizes the experimental results on the Shanghai and Nanchang datasets. The results indicate that LLMApp outperforms all baseline models on all evaluation metrics across both datasets. Furthermore, the results highlight several key observations:

\textbf{LLMApp better captures the dependencies between spatiotemporal context and app category}. LLMApp demonstrates noticeably better performance than CADHG and MAPLE(e.g., LLMApp MRR@1 = 0.6382 vs. CADHG = 0.3980 and MAPLE = 0.5850), and this is because both MAPLE and CADHG only treat spatiotemporal context as auxiliary inputs without fully exploring the dependencies between location, time, and app category—instead, LLMApp leverages LLMs' contextual reasoning to learn these relationships more deeply.
Using a coarse-to-fine hierarchical architecture, LLMApp first forecasts the app category based on location and time, and then predicts the specific app using this intermediate signal. This strategy leads to clear improvements in prediction accuracy and interpretability.

\textbf{LLMApp better models user habits from long-term app usage record sequence}. Compared with MAPLE, LLMApp's stronger predictive performance (e.g., LLMApp ACC@1 = 0.6382 vs. MAPLE ACC@1 = 0.5850) also stems from extracting user habits from the long-term app usage record sequence.
Specifically, when selecting among apps within the same category or under similar contextual conditions, LLMApp can leverage refined user profiles to select the app that best matches the user's preferences, leading to more accurate predictions. Compared with other user profile modeling methods, such as AppUsage2Vec and DUGN, LLMApp selects informative records from redundant histories, retaining only those that best reflect user preferences and behavioral patterns. This selection process removes noise and strengthens the quality of the remaining data.

\textbf{LLMApp better encodes diverse contextual signals into semantic representations}. Compared with graph-based approaches (e.g., DUGN, SA-GCN) and hypergraph-based methods (e.g., SHARE, CADHG), LLMApp leverages LLMs to capture relationships among contextual features more effectively  (e.g., LLMApp NDCG@5 = 0.6382 vs. DUGN = 0.3980 and CADHG = 0.5850), since LLMs offer stronger context understanding, more profound basic knowledge of human behavior patterns, and more capable reasoning skills. These advantages generally lead to higher prediction accuracy.

\begin{table*}[tb]
\centering
\footnotesize
\caption{Evaluation results of different LLM backbones of LLMApp on two real-world datasets.}
\label{tab:llm_backbone}
\resizebox{0.85\linewidth}{!}
{
\begin{tabular}{p{2.3cm}|ccccccccc}
\toprule
\multicolumn{1}{c|}{\textbf{Dataset}} & \multicolumn{9}{c}{\textbf{Shanghai}} \\
\midrule
\multirow{2}{*}{\raisebox{-0.3\height}{\textbf{LLM Backbones}}}
  & \multicolumn{3}{c}{\textbf{Acc@K}}
  & \multicolumn{3}{c}{\textbf{MRR@K}}
  & \multicolumn{3}{c}{\textbf{NDCG@K}} \\
  \cmidrule(lr){2-4}\cmidrule(lr){5-7}\cmidrule(lr){8-10}
  & 1 & 5 & 10 & 1 & 5 & 10 & 1 & 5 & 10 \\
\midrule
\textcolor{black}{MobileLLM-600M} & 0.5536 & 0.8214 & 0.8732 & 0.5536 & 0.6178 & 0.6315 & 0.5536 & 0.7043 & 0.7398 \\
\textcolor{black}{MobileLLM-1B} & 0.5712 & 0.8369 & 0.8835 & 0.5712 & 0.6264 & 0.6397 & 0.5712 & 0.7162 & 0.7486 \\
\textcolor{black}{Gemma-2-2B} & 0.5789 & 0.8427 & 0.8869 & 0.5789 & 0.6298 & 0.6435 & 0.5789 & 0.7207 & 0.7525 \\
\textcolor{black}{Gemma-2-9B} & 0.6028 & 0.9085 & 0.9458 & 0.6028 & 0.6843 & 0.6951 & 0.6028 & 0.7656 & 0.8324 \\
Llama-2-7B & 0.6382 & 0.9306 & 0.9480
& 0.6382 & 0.7084 & 0.7153
& 0.6382 & 0.8017 & 0.8485 \\
Llama-2-13B & \textbf{0.6953} & \textbf{0.9313} & \textbf{0.9500} & \textbf{0.6953} & \textbf{0.7492} & \textbf{0.7600} & \textbf{0.6953} & \textbf{0.8692} & \textbf{0.9152} \\
DeepSeek-R1-1.5B & 0.5189 & 0.7437 & 0.7881 & 0.5189 & 0.5786 & 0.6062 & 0.5189 & 0.6426 & 0.6605 \\
DeepSeek-R1-7B & 0.6316 & 0.8895 & 0.8947 & 0.6316 & 0.6947 & 0.7075 & 0.6316 & 0.7984 & 0.8311 \\
\midrule

\multicolumn{1}{c|}{\textbf{Dataset}} & \multicolumn{9}{c}{\textbf{Nanchang}} \\
\midrule
\multirow{2}{*}{\raisebox{-0.3\height}{\textbf{LLM Backbones}}}
  & \multicolumn{3}{c}{\textbf{Acc@K}}
  & \multicolumn{3}{c}{\textbf{MRR@K}}
  & \multicolumn{3}{c}{\textbf{NDCG@K}} \\
  \cmidrule(lr){2-4}\cmidrule(lr){5-7}\cmidrule(lr){8-10}
  & 1 & 5 & 10 & 1 & 5 & 10 & 1 & 5 & 10 \\
\midrule
\textcolor{black}{MobileLLM-600M} & 0.4628 & 0.7824 & 0.8846 & 0.4628 & 0.5417 & 0.5685 & 0.4628 & 0.5976 & 0.6258 \\
\textcolor{black}{MobileLLM-1B} & 0.4765 & 0.7988 & 0.8967 & 0.4765 & 0.5532 & 0.5794 & 0.4765 & 0.6094 & 0.6371 \\
\textcolor{black}{Gemma-2-2B} & 0.4809 & 0.8065 & 0.9032 & 0.4809 & 0.5575 & 0.5848 & 0.4809 & 0.6145 & 0.6433 \\
\textcolor{black}{Gemma-2-9B} & 0.5085 & 0.8286 & 0.9236 & 0.5085 & 0.5681 & 0.5740 & 0.5085 & 0.6206 & 0.6683 \\
Llama-2-7B & 0.5040 & 0.8810 & 0.9350 & 0.5040 & 0.5923 & 0.5936 & 0.5040 & 0.6446 & 0.6819 \\
Llama-2-13B & \textbf{0.5472} & \textbf{0.8866} & \textbf{0.9385} & \textbf{0.5472} & \textbf{0.6078} & \textbf{0.6418} & \textbf{0.5472} & \textbf{0.7841} & \textbf{0.7470}\\
DeepSeek-R1-1.5B & 0.4588 & 0.7178 & 0.8336 & 0.4588 & 0.5545 & 0.5702 & 0.4588 & 0.5952 & 0.6328 \\
DeepSeek-R1-7B & 0.4843 & 0.8376 & 0.9124 & 0.4843 & 0.5770 & 0.5922 & 0.4843 & 0.6172 & 0.6534 \\
\bottomrule
\end{tabular}}
\end{table*}

\begin{figure}[tb]
\vspace{-3mm}
\centering
\includegraphics[width=1.0\linewidth]{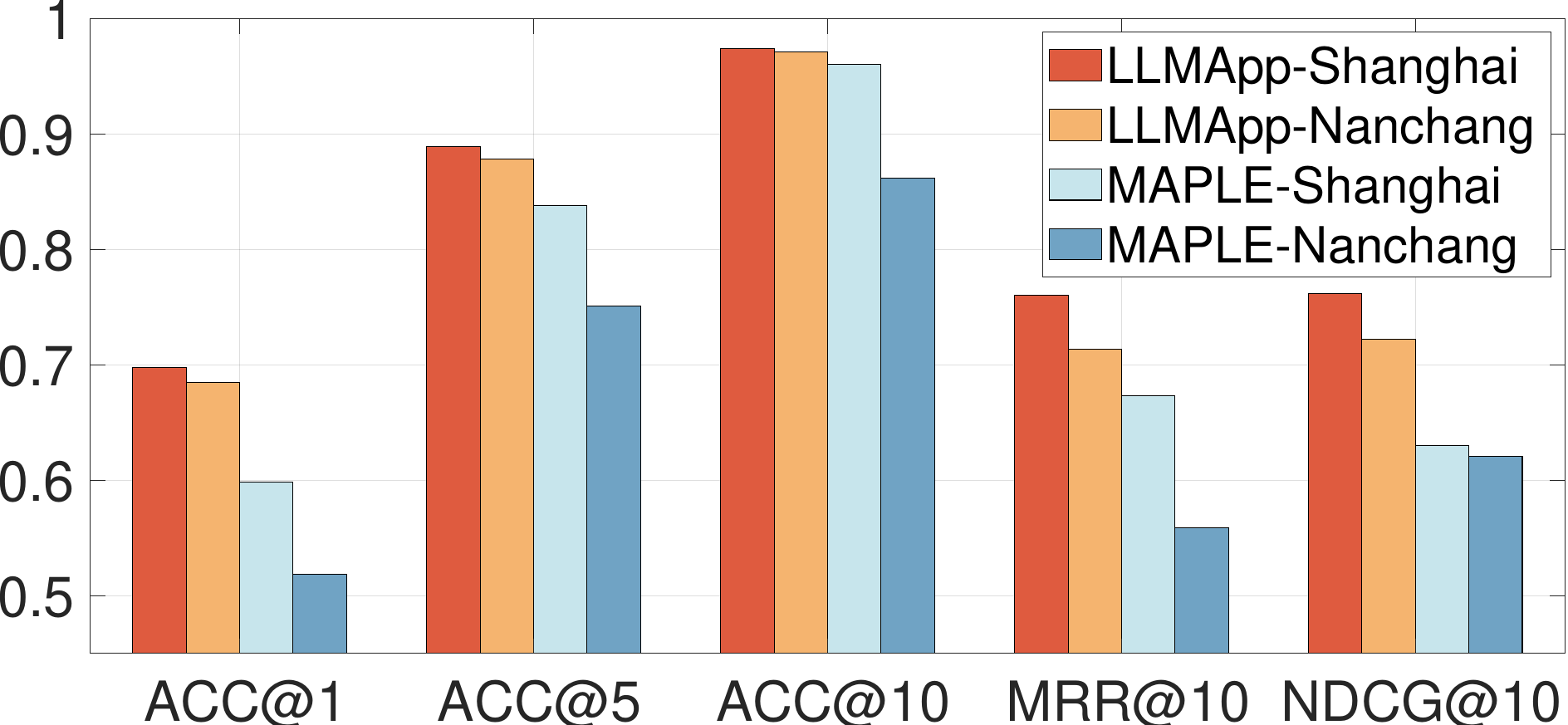}
\vspace{-5mm}
\caption{Evaluation results of app category prediction accuracy.}
\label{fig:cate_accuracy}
\vspace{-3mm}
\end{figure}

\begin{figure*}[tb]
\centering
\includegraphics[width=1.0\linewidth]{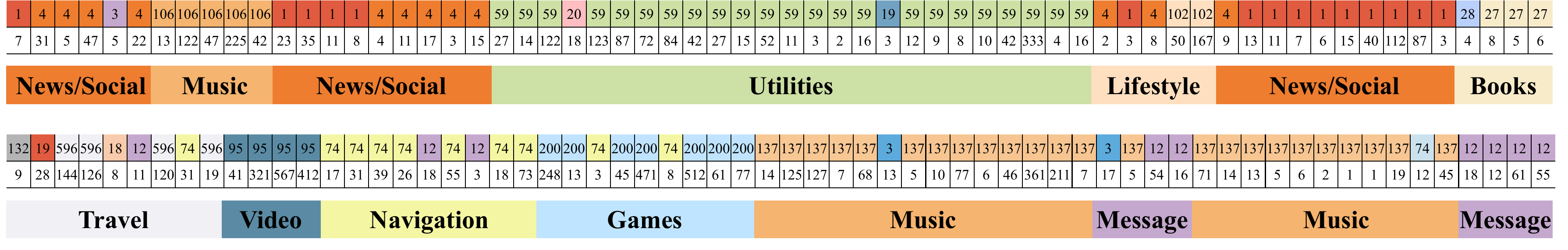}
\vspace{-2mm}
\caption{Visualization of data compression. The upper part demonstrates raw long-term records from the Shanghai and Nanchang datasets, where each record contains app ID (the first row), usage time (the second row), and app category with colors. The lower part is compressed usage records.}
\label{fig:case_compression}
\vspace{-2mm}
\end{figure*}

\subsubsection{Ablation Study}
To examine the specific contributions of each component within LLMApp, we conducted systematic ablation experiments by removing key components and evaluating their impact on mobile app usage prediction. Three ablation variants were designed: (1) \textbf{w/o hierarchical} predicts the app usage based on contextual signals and historical sequences, without first estimating the app category; (2) \textbf{w/o app category} omits the dependencies between spatiotemporal context and app category, and relies only on user habits and historical sequences for prediction; (3) \textbf{w/o user habits} predicts app usage based solely on spatiotemporal context and the inferred app category, without modeling user habits.
Table~\ref{tab:ablation} reveals that removing any component leads to notable performance degradation across critical metrics, underscoring their distinct but complementary roles. 
The w/o user habits variant shows an apparent decrease across all metrics, especially in ACC@1 (0.5767 vs. 0.6382) and ACC@5 (0.8177 vs. 0.9306), indicating that although the variant can still estimate the app category, lacking knowledge of the user’s long-term preferences and usage tendencies makes it challenging to select the correct app within the predicted category. 
The w/o app category variant also suffers substantial performance degradation across all metrics (e.g., ACC@5 0.7307 vs. 0.9306, MRR@5 0.5781 vs. 0.7084), confirming that enabling the LLM to model the dependencies between app category and spatiotemporal context is crucial for accurate app usage prediction. The coarse-to-fine structure significantly narrows the candidate space. 
The w/o hierarchical variant yields the most severe performance drop. Without the hierarchical structure, the model fails to effectively incorporate contextual signals and user profile information, resulting in significant declines across all metrics. These results show that, even with LLMs' strong contextual understanding and semantic reasoning abilities, accurate app usage prediction requires proper designs for processing and extracting contextual signals.

\begin{figure*}[tb]
\centering
\begin{subfigure}{0.22\linewidth}
    \centering
    \includegraphics[width=\linewidth]{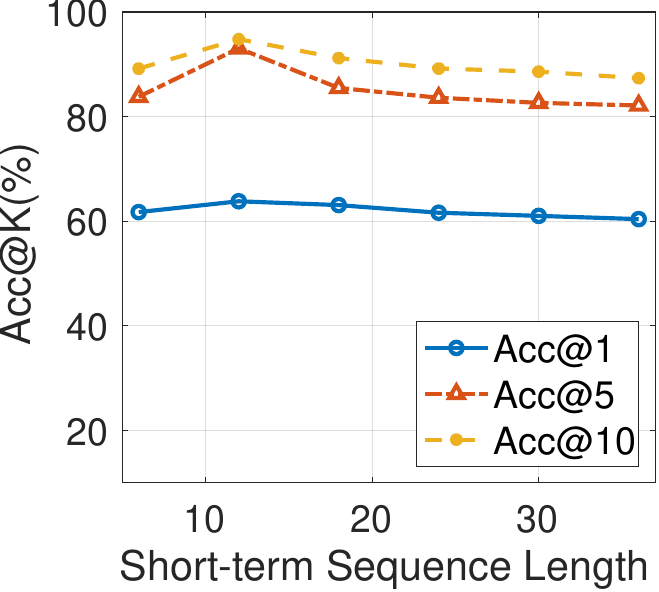}
    \caption{ACC@K on Shanghai}
\end{subfigure}
\begin{subfigure}{0.22\linewidth}
    \centering
    \includegraphics[width=\linewidth]{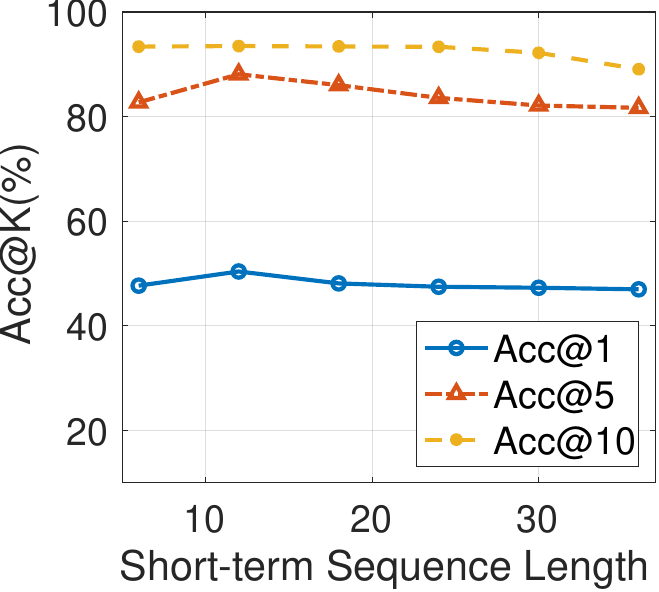}
    \caption{ACC@K on Nanchang}
\end{subfigure}
\begin{subfigure}{0.22\linewidth}
    \centering
    \includegraphics[width=\linewidth]{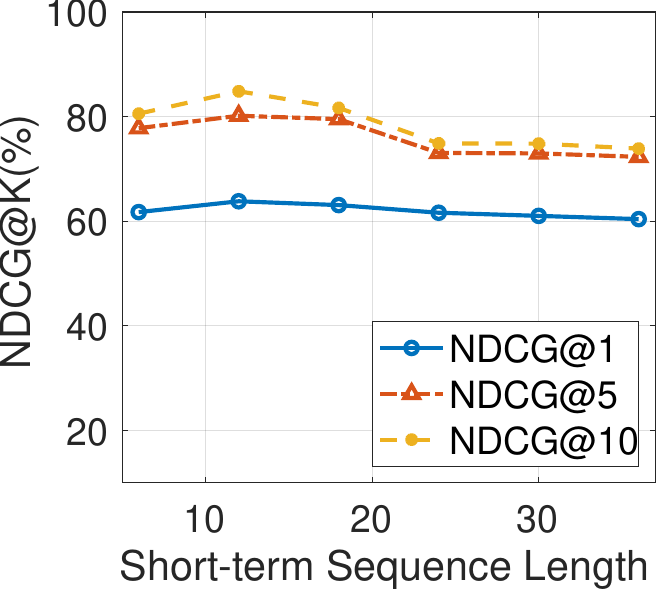}
    \caption{NDCG@K on Shanghai}
\end{subfigure}
\begin{subfigure}{0.22\linewidth}
    \centering
    \includegraphics[width=\linewidth]{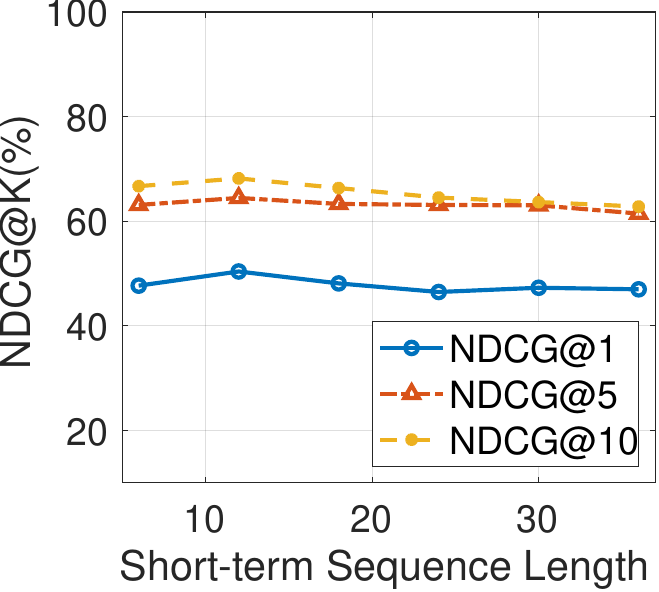}
    \caption{NDCG@K on Nanchang}
\end{subfigure}
\caption{Sensitivity of short-term sequence length on LLMApp's performance across Shanghai and Nanchang datasets.}
\label{fig:Hyperparameter_short}
\vspace{-3mm}
\end{figure*}

\begin{figure*}[tb]
\centering
\begin{subfigure}{0.22\linewidth}
    \centering
    \includegraphics[width=\linewidth]{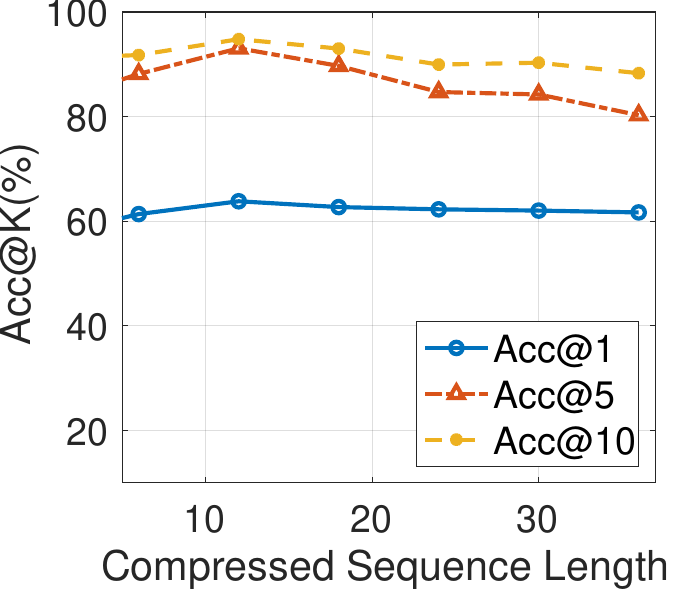}
    \caption{ACC@K on Shanghai}
\end{subfigure}
\begin{subfigure}{0.22\linewidth}
    \centering
    \includegraphics[width=\linewidth]{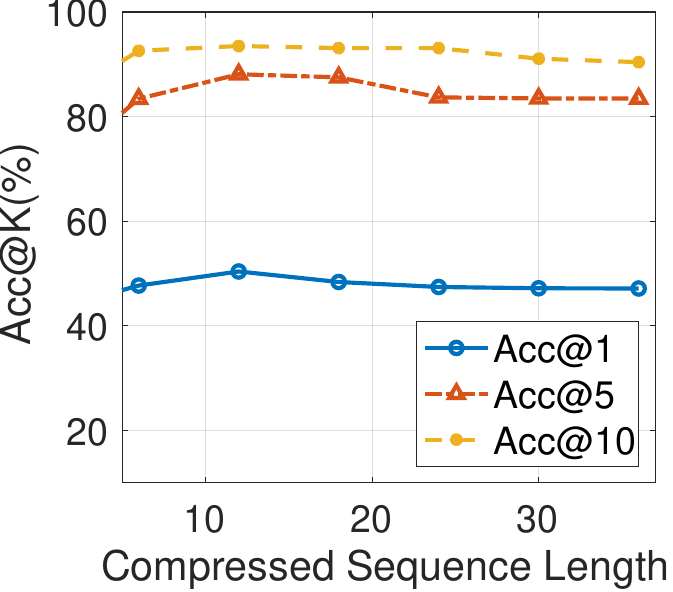}
    \caption{ACC@K on Nanchang}
\end{subfigure}
\begin{subfigure}{0.22\linewidth}
    \centering
    \includegraphics[width=\linewidth]{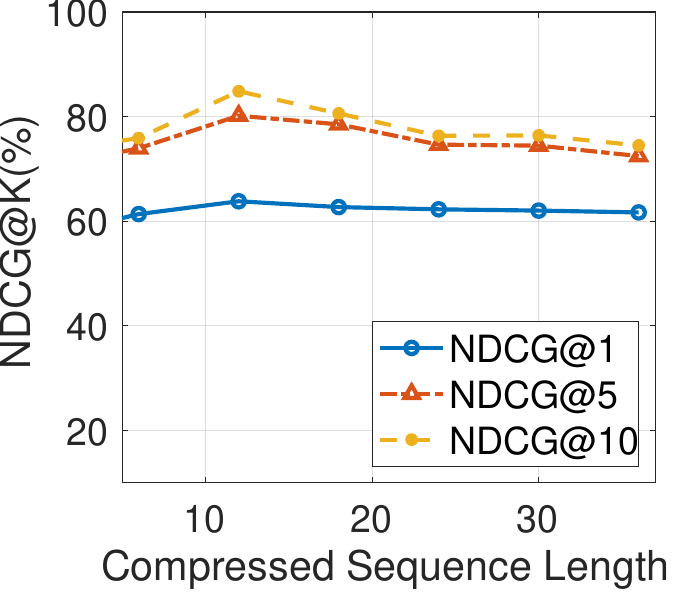}
    \caption{NDCG@K on Shanghai}
\end{subfigure}
\begin{subfigure}{0.22\linewidth}
    \centering
    \includegraphics[width=\linewidth]{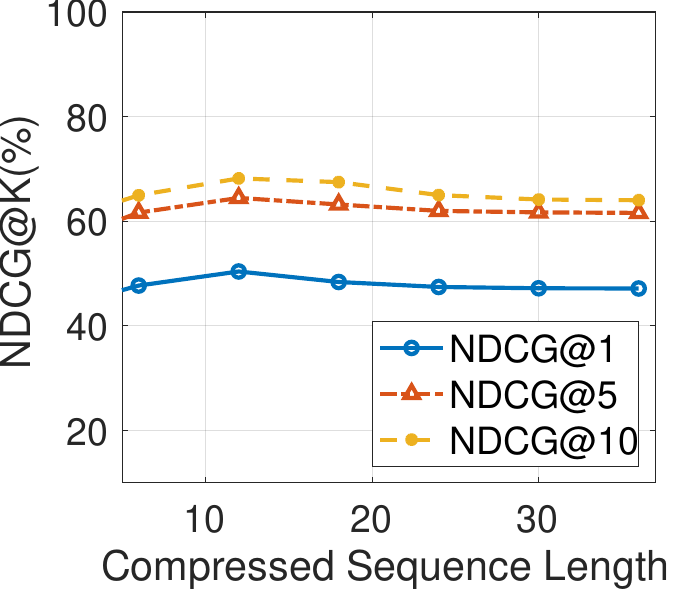}
    \caption{NDCG@K on Nanchang}
\end{subfigure}
\caption{\textcolor{black}{Sensitivity of compressed sequence length on LLMApp's performance across Shanghai and Nanchang datasets.}}
\label{fig:Hyperparameter_long}
\vspace{-3mm}
\end{figure*}

\subsubsection{App Category Prediction Accuracy}
We conducted a comparative evaluation of prediction accuracy on the Shanghai and Nanchang datasets to assess the effectiveness of LLMApp in capturing dependencies between spatiotemporal context and app category usage patterns. 
As illustrated in Fig.~\ref{fig:cate_accuracy}, compared with MAPLE, LLMApp consistently achieves higher accuracy across all metrics, demonstrating its effectiveness in capturing the underlying dependencies between spatiotemporal context and app category usage and in accurately forecasting category-level behaviors.
Compared with MAPLE, LLMApp exhibits notably stronger performance on ACC@5 and ACC@10, reflecting its enhanced capability to interpret contextual signals and produce reliable category predictions. The improvements in NDCG@10 further indicate that LLMApp not only identifies the correct category but also ranks relevant categories more appropriately. Overall, the results confirm that LLMApp provides a more precise, context-aware model of mobile app usage patterns across diverse spatiotemporal conditions.

\subsubsection{Data Compression Case Study}
We conduct a case study to evaluate LLMApp’s ability to extract user habits from long-term app usage sequences. As illustrated in Fig.~\ref{fig:case_compression}, we visualized the results before and after compression. Before compression, records are highly redundant, noisy, and fragmented, containing many short, semantically weak operations, which makes it difficult to understand the user’s actual behavioral patterns.
After processing with the long-term data compressor, the lower part of Fig.~\ref{fig:case_compression} shows the compressed usage records. Compared with the original sequence, the compressed version yields much more precise and stable semantic segments (e.g., News/Social, Music, Utilities, Games, Messaging), which better reveal the user’s dominant behavioral tendencies across different time points. For example, scattered initially social app usage consolidates into coherent segments, while high-frequency but meaningless short switches are filtered out.
This case study demonstrates that the long-term data compressor effectively removes noise and retains the most informative behavioral patterns, producing a compact yet semantically complete representation of the user’s long-term preferences. Such compressed records not only better reflect real usage habits but also provide higher-quality signals for downstream LLM reasoning, significantly improving prediction accuracy and reliability.

\subsubsection{LLM Backbone Comparison}
\textcolor{black}{We conduct an LLM backbone comparison to evaluate how different language models influence LLMApp's performance. We compare eight backbones with different parameter scales and architectures, including MobileLLM-600M, MobileLLM-1B, Gemma-2-2B, Gemma-2-9B, DeepSeek-R1-Distill-Qwen-1.5B, DeepSeek-R1-Distill-Qwen-7B, LLaMA-2-7B, and LLaMA-2-13B. Among them, MobileLLM-600M and MobileLLM-1B are included as lightweight mobile-oriented backbones to examine whether LLMApp can work with smaller language models.
As shown in Table~\ref{tab:llm_backbone}, larger backbones generally provide stronger semantic reasoning ability and achieve better prediction performance. LLaMA-2-13B obtains the best overall results, while LLaMA-2-7B and DeepSeek-R1-Distill-Qwen-7B also perform competitively. Meanwhile, MobileLLM-600M and MobileLLM-1B still achieve reasonable performance despite their much smaller parameter scales, indicating that LLMApp is not restricted to large LLMs and can be adapted to lightweight backbones for resource-constrained scenarios.}

\subsubsection{Hyperparameter Study}
\label{sensitivity}
We conduct a hyperparameter study to examine how the lengths of the short-term and compressed long-term app usage record sequences affect model performance.
Fig.~\ref{fig:Hyperparameter_short} and Fig.~\ref{fig:Hyperparameter_long} reports the results across Acc@K, MRR@K, and NDCG@K on Shanghai and Nanchang datasets.
For short-term sequence length, increasing the window from 6 to around 12 generally improves performance, as slightly longer context helps the model capture recent behavioral intent. Beyond this point, further extending the window yields diminishing returns or mild degradation, likely due to the inclusion of outdated or noisy usage events.
\textcolor{black}{For the length of the compressed sequence, performance remains consistently strong across a broad range, showing only slight variation as the sequence lengthens. This indicates that the compressor can extract compact yet informative user habits, and that the model is robust to the specific choice of pattern length, providing a practical trade-off between accuracy and efficiency.}
Overall, the results demonstrate that LLMApp is stable across a wide range of hyperparameter settings, and both short-term and long-term components effectively preserve proper behavioral signals without being overly sensitive to length choices.

\subsubsection{Generalization} 

\begin{figure}[tb]
\centering
\vspace{-2mm}
\begin{subfigure}{0.45\linewidth}
    \centering
    \includegraphics[width=\linewidth]{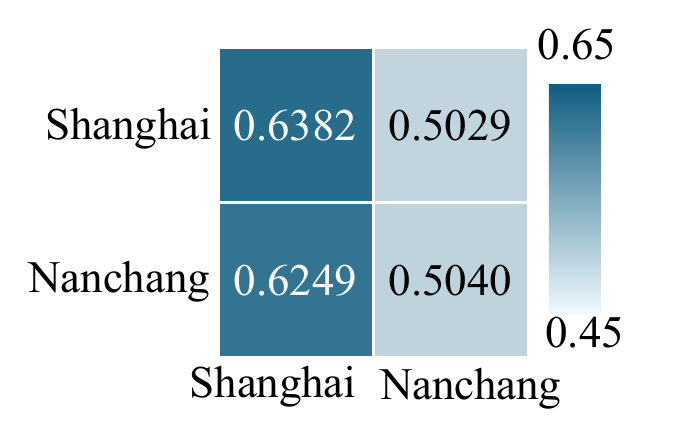}
    \caption{ACC@1}
\end{subfigure}
\begin{subfigure}{0.45\linewidth}
    \centering
    \includegraphics[width=\linewidth]{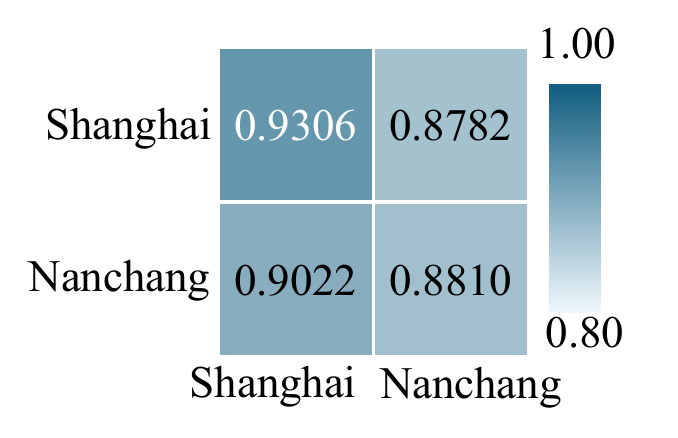}
    \caption{ACC@5}
\end{subfigure}

\begin{subfigure}{0.45\linewidth}
    \centering
    \includegraphics[width=\linewidth]{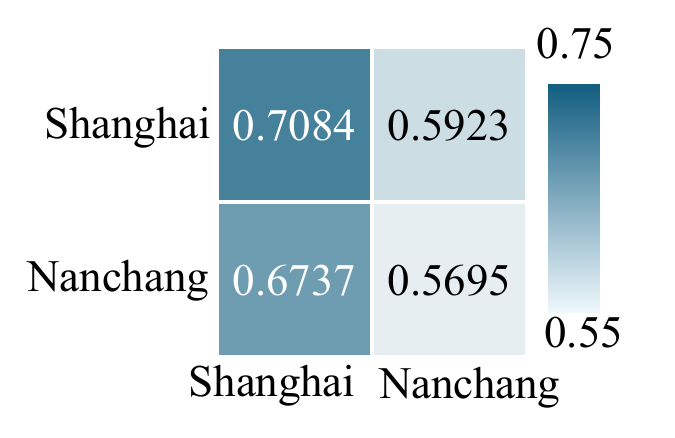}
    \caption{MRR@5}
\end{subfigure}
\begin{subfigure}{0.45\linewidth}
    \centering
    \includegraphics[width=\linewidth]{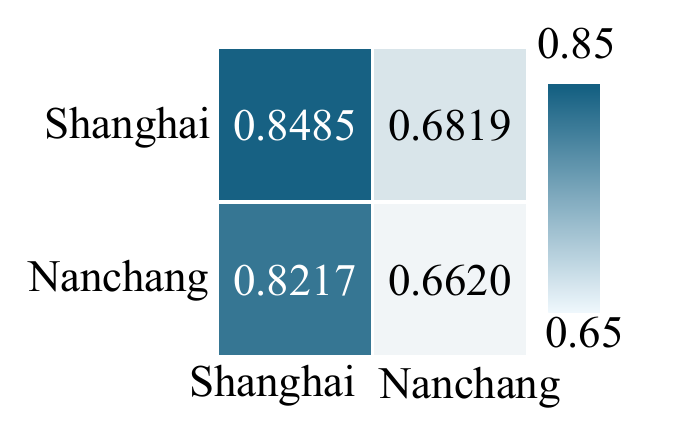}
    \caption{NDCG@10}
\end{subfigure}
\vspace{-2mm}
\caption{Generalization experiments on Shanghai and Nanchang datasets. Darker grids denote better generalization and robustness.}
\label{fig:Generalization}
\vspace{-2mm}
\end{figure}

We conducted experiments to evaluate the generalization and robustness of LLMApp. We conducted cross-city experiments in which models trained in one city were applied to predict app usage in another city. Specifically, we train our model on the Shanghai dataset and deploy it to predict app usage on the Nanchang dataset, and vice versa.
This experiment assesses LLMApp's adaptability across different urban environments. The results in Fig.~\ref{fig:Generalization} demonstrate that LLMApp maintains high prediction accuracy and reliability, exhibiting strong adaptability. 

\subsubsection{Spatial Dependencies Modeling Quality}

\begin{figure}[tb]
\centering
\vspace{-2mm}
\begin{subfigure}{0.49\linewidth}
    \centering
    \includegraphics[width=\linewidth]{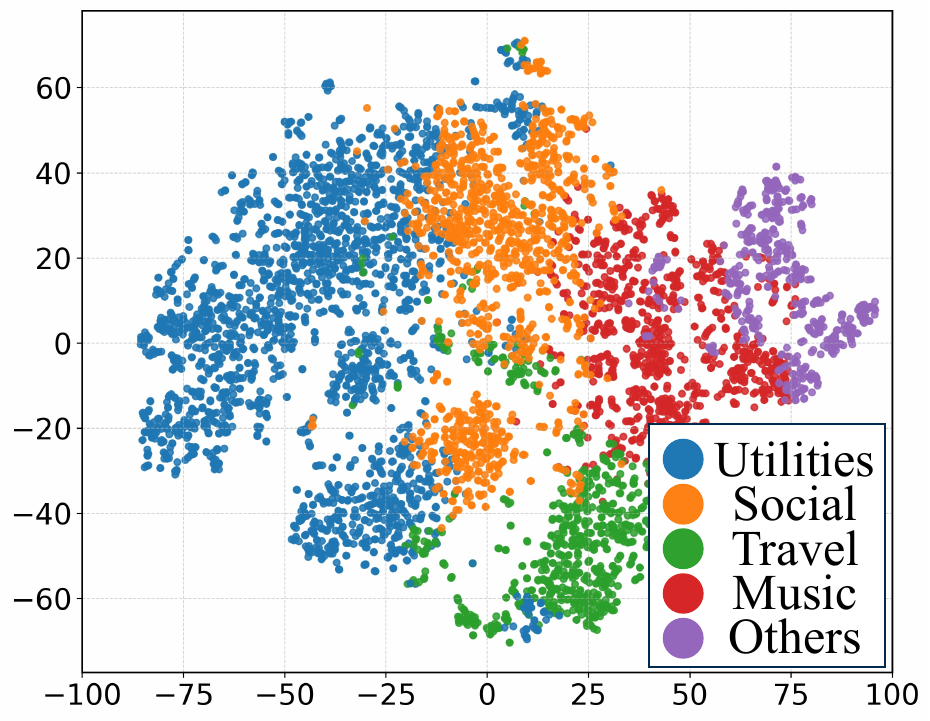}
\end{subfigure}
\begin{subfigure}{0.49\linewidth}
    \centering
    \includegraphics[width=\linewidth]{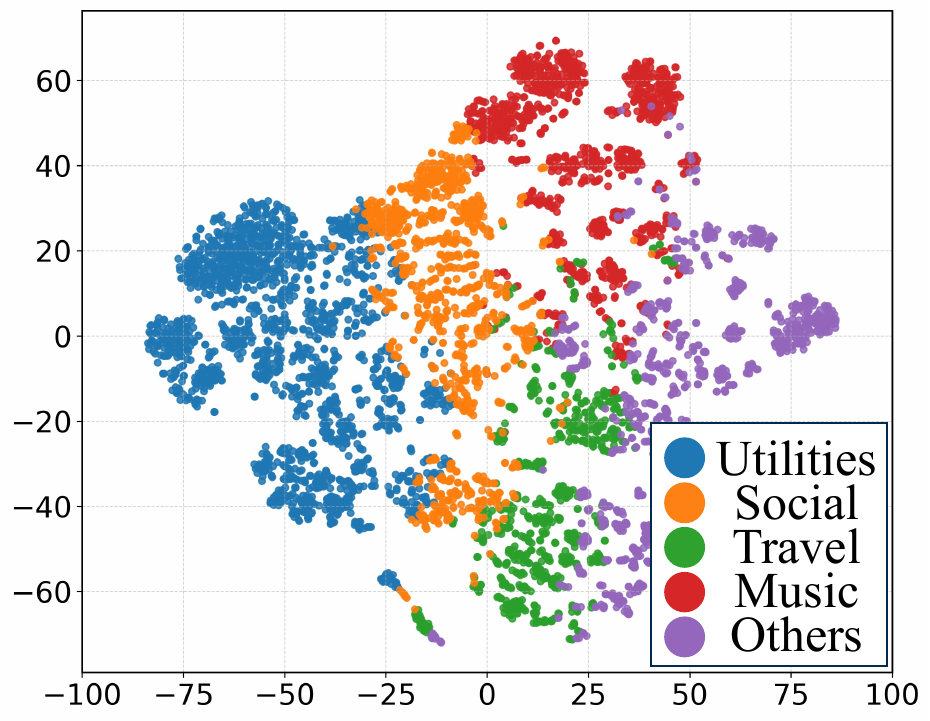}
\end{subfigure}
\caption{Visualization of location embeddings learned by MAPLE (left) and LLMApp (right)}
\label{fig:tsne}
\end{figure}

To evaluate the effectiveness of LLMApp in modeling spatial dependencies in app usage, we conducted a location-based cluster analysis. 
Each location is represented by a contextual embedding learned by a predictive model, which is then projected into a two-dimensional space for visualization via t-SNE~\cite{maaten2008visualizing}.
Next, each location is labeled using K-means clustering based on its empirical app-category distribution.
As shown in Fig.~\ref{fig:tsne}, LLMApp yields more coherent, semantically aligned spatial clusters than MAPLE, demonstrating stronger capability to capture spatial dependencies in app usage behavior.

\subsubsection{Traffic Prediction Performance}

We conduct experiments on two real-world datasets and compare LLMApp against a set of representative baselines to evaluate its effectiveness for traffic volume prediction. The evaluation follows the standard regression setting and measures prediction performance using MAE, MSE, RMSE, and Spearman’s correlation.
We include the following baseline models: Naive, Linear Regression, LSTM, TCN, Transformer, CADHG, and MAPLE.
As shown in Table~\ref{tab:traffic_prediction}, LLMApp consistently achieves the lowest MAE, MSE, and RMSE and the highest Spearman correlation across both datasets. Traditional sequence models (LSTM, TCN, Transformer) capture short-term fluctuations but struggle with long-range dependencies and the nonlinear influence of contextual factors. Hypergraph-based CADHG and LLM-based MAPLE improve performance by modeling richer structural information, yet they remain limited in encoding diverse contextual signals and extracting user-specific traffic patterns. In contrast, LLMApp integrates hierarchical context modeling and user habit extraction, enabling accurate estimation of traffic magnitude even under noisy, fragmented usage patterns.
Overall, the results demonstrate that LLMApp delivers the most reliable and stable traffic prediction among all baselines, highlighting the advantage of unified contextual encoding and long-term habit modeling for fine-grained traffic estimation.

\begin{table}[tb]
\centering
\caption{Traffic prediction performance of different models.}
\label{tab:traffic_prediction}
\vspace{-2mm}
\footnotesize
\resizebox{0.8\linewidth}{!}{
\begin{tabular}{c|cccc}
\toprule
\multirow{2}{*}{\raisebox{-0.3\height}{\textbf{Methods}}}
    & \multicolumn{4}{c}{\textbf{Shanghai}} \\
    \cmidrule(lr){2-5}
    & \textbf{MAE} & \textbf{MSE} & \textbf{RMSE} & \textbf{Spearman} \\
\midrule
Naive        & 15.84 & 1385  & 49.22 & 0.2793 \\
Linear R     & 11.15 & 896.7 & 37.94 & 0.3797 \\
LSTM         & 10.96 & 905.0 & 30.83 & 0.3865 \\
TCN          & 10.99 & 908.0 & 30.13 & 0.3812 \\
Transformer  & 10.94 & 919.6 & 30.10 & 0.3815 \\
HyperG       & 9.716 & 549.0 & 21.06 & 0.4011 \\
MAPLE        & 8.101 & 413.9 & 19.55 & 0.4590 \\
\textbf{LLMApp} & \textbf{7.253} & \textbf{352.3} & \textbf{18.77} & \textbf{0.4710} \\
\midrule
\multirow{2}{*}{\textbf{Methods}} 
    & \multicolumn{4}{c}{\textbf{Shanghai}} \\
    \cmidrule(lr){2-5}
    & \textbf{MAE} & \textbf{MSE} & \textbf{RMSE} & \textbf{Spearman} \\
\midrule
Naive        & 15.27 & 3312 & 57.55 & 0.3047 \\
Linear R     & 11.78 & 2525 & 55.00 & 0.3649 \\
LSTM         & 9.565 & 2054 & 45.32 & 0.3924 \\
TCN          & 9.581 & 2043 & 45.21 & 0.3870 \\
Transformer  & 9.569 & 2058 & 45.37 & 0.4056 \\
CADHG       & 9.003 & 1461 & 29.96 & 0.4162 \\
MAPLE        & 8.951 & 1277 & 26.15 & 0.4333 \\
\textbf{LLMApp} & \textbf{8.594} & \textbf{1079} & \textbf{23.78} & \textbf{0.4596} \\
\bottomrule
\end{tabular}
}
\vspace{-2mm}
\end{table}

\begin{figure*}[tb]
\vspace{-2mm}
\centering
\includegraphics[width=0.9\textwidth]{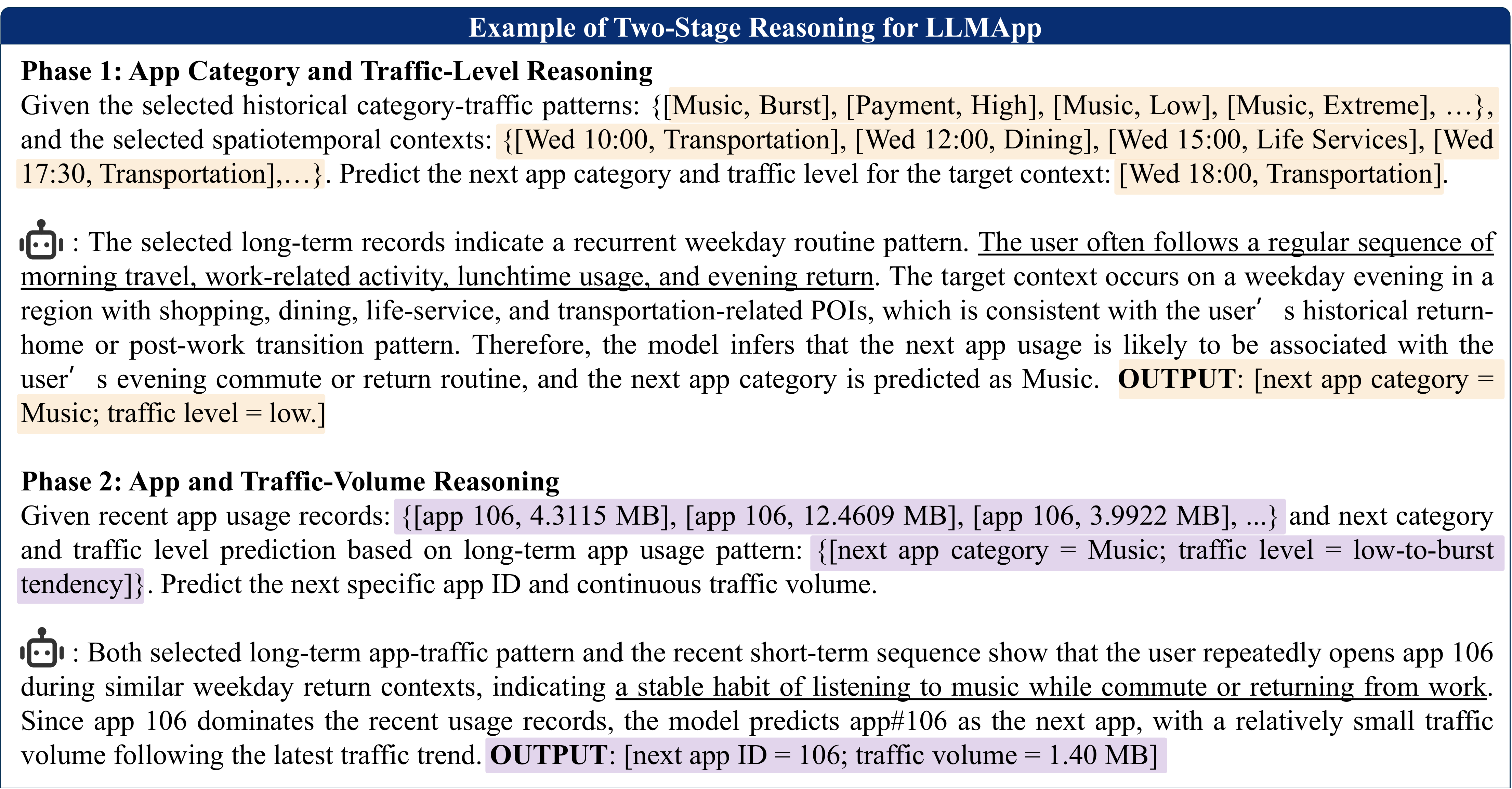}
\vspace{-1mm}
\caption{\textcolor{black}{Interpretability case study of the two-stage reasoning process in LLMApp.}}
\label{fig:interpretability}
\vspace{-2mm}
\end{figure*}

\subsubsection{Complexity Analysis}
\textcolor{black}{
To evaluate the practical feasibility of LLMApp, we compare its runtime, token consumption, estimated energy consumption, and prediction accuracy against MAPLE on the Shanghai dataset, with both methods using Gemma-2-2B as the LLM backbone. As shown in Table~\ref{tab:complexity}, LLMApp achieves lower inference latency and consumes fewer tokens and energy than MAPLE. This advantage primarily stems from the prompt compressor, which filters out redundant long-term app usage records and retains only compact key patterns for downstream LLM reasoning. Therefore, LLMApp reduces both the input token cost and the communication overhead caused by transmitting long textual prompts. These results indicate that LLMApp not only improves prediction accuracy, but also supports the feasibility of edge-side inference in practical deployment scenarios.}
\textcolor{black}{
Moreover, the efficiency of LLMApp can be further improved by adjusting the compressed sequence length of long-term memory. As the sensitivity analysis of compressed sequence length shown in Sec.~\ref{sensitivity} indicates, LLMApp can still maintain competitive prediction accuracy with a shorter compressed sequence, thereby further reducing token consumption, memory usage, and inference overhead. This indicates that the compressed sequence length can be flexibly selected according to the resource budget of different deployment scenarios. In addition, as efficient LLM inference, model compression, and edge computing hardware continue to develop, LLMApp can be more easily integrated into edge-assisted mobile service systems in the future.}

\subsubsection{Interpretability Analysis}

\begin{table}[t]
\centering
\caption{\textcolor{black}{Complexity comparison on the Shanghai dataset.}}
\label{tab:complexity}
\footnotesize
\setlength{\tabcolsep}{3.5pt}
\renewcommand{\arraystretch}{1.08}
\begin{tabular}{lccccc}
\toprule
\textbf{Model} & \textbf{Train} & \textbf{Infer} & \textbf{Tokens} & \textbf{Energy}  & \textbf{ACC@1} \\
& \textbf{(s/sample)}  & \textbf{(s/sample)} & \textbf{(/sample)} & \textbf{(J/sample)} & \\
\midrule
MAPLE  & 0.0189 & 0.0096 & 1080 & 1.4400 & 0.5471 \\
LLMApp & 0.0173 & 0.0071 & 942 & 1.0650 & 0.5789 \\
\bottomrule
\end{tabular}
\end{table}

\textcolor{black}{
To better explain how LLMApp connects contextual evidence with prediction results, we provide an interpretability case study. Since LLMApp performs prediction using soft-prefix embeddings rather than direct text generation, the reasoning shown in the figure is a human-readable interpretation of the underlying process.
As shown in Fig.~\ref{fig:interpretability}, LLMApp provides a two-stage interpretable reasoning case. In the first stage, the selected category-traffic patterns and spatiotemporal contexts reveal a recurrent weekday routine, such as work-related activity, lunchtime usage, and evening return, leading the model to predict Music as the next app category. In the second stage, LLMApp combines this category-level signal with long-term and recent app-traffic records. Since app 106 frequently appears in similar weekday return contexts, the model predicts app 106 with a relatively small traffic volume, showing how spatiotemporal routines and user habits support the final prediction.
}

\subsubsection{Prompt Variation Analysis}
\begin{table}[t]
\centering
\caption{\textcolor{black}{Prompt robustness analysis on the Shanghai dataset.}}
\label{tab:prompt_robustness}
\footnotesize
\setlength{\tabcolsep}{4pt}
\begin{tabular}{lcccc}
\toprule
\textbf{Prompt Variant} & \textbf{ACC@1} & \textbf{ACC@5} & \textbf{MRR@5} & \textbf{NDCG@10} \\
\midrule
Original prompt & 0.6382 & 0.9306 & 0.7084 & 0.8485 \\
Simplified prompt & 0.6315 & 0.9248 & 0.7080 & 0.8421 \\
\bottomrule
\end{tabular}
\end{table}

\textcolor{black}{
To examine whether LLMApp is sensitive to prompts, we conduct a prompt-variation analysis on the Shanghai dataset. Specifically, we compare two prompt settings: the original prompt used in LLMApp, which describes app usage and spatiotemporal contexts in natural language, and a simplified prompt that retains only structured key fields such as app category, traffic level, time, and POI information. The results show that the simplified prompt yields only slight performance changes relative to the original prompt, indicating that LLMApp is not sensitive to specific handcrafted natural-language templates.
}

%% file: Related.tex
\section{Related Work}
\label{sec:related}

\subsection{Mobile App Usage Prediction}
Early efforts in mobile app usage prediction primarily employed sequential models such as Long Short-Term Memory (LSTM)~\cite{hochreiter1997long} networks to capture temporal dependencies in user behavior. These methods~\cite{katsarou2022whatsnextapp,xu2018predicting,cheng2023mobile} aimed to learn latent user intent by analyzing historical app usage sequences.
To incorporate contextual information, researchers extended these models with personalized or structure-aware mechanisms~\cite{huang2012predicting,cao2017mining,baeza2015predicting,frey2017mobile}. Zhao~\emph{et al.}~\cite{zhao2019appusage2vec} proposed AppUsage2Vec, which leverages an attention-based framework and dual DNN modules to learn user-specific app usage representations by measuring each app’s contribution to the target app. Chen~\emph{et al.}~\cite{chen2019cap} introduced CAP, a context-aware prediction method that constructs heterogeneous graphs linking apps with contextual signals such as time, location, and app type. Yu~\emph{et al.}~\cite{yu2020semantic} designed SA-GCN, which formulates app usage data as a graph and applies graph convolution to capture semantic dependencies.
Graph-based modeling was further explored in DUGN~\cite{ouyang2022learning}, proposed by Ouyang~\emph{et al.}, which dynamically constructs app usage graphs to model evolving user intent. SHARE~\cite{wang2021session}, introduced by Wang~\emph{et al.}, encodes session-level context using hypergraph attention mechanisms. CADHG~\cite{huang2025predicting}, proposed by Huang~\emph{et al.}, builds context-aware dynamic hypergraphs that fuse user trajectories with app sessions to model higher-order spatiotemporal relations.
Recently, Khaokaew~\emph{et al.}~\cite{khaokaew2024maple} presented MAPLE, which employs large language models to embed contextual data (e.g., location, time, installed apps) into a unified semantic representation. MAPLE adopts a prompt-based framework and performs hierarchical app usage prediction, demonstrating promising results, especially under cold-start conditions.

\subsection{Large Language Models}
Recent advances show that large language models can be reprogrammed for behavioral and spatiotemporal prediction by converting structured signals into natural-language style prompts. LLM-based recommenders~\cite{sun2019bert4rec,lyu2024llm,li2023gpt4rec} treat interaction histories as text and leverage semantic reasoning for sequential prediction~\cite{lin2024rella,li2023text}, while Time-LLM~\cite{jintime} and related works adapt frozen LLMs to time-series forecasting through prefix-style conditioning. In mobility analysis, POI-LLM~\cite{kim2025large}, Mobility-LLM~\cite{gong2024mobility}, and LingoTrip~\cite{qin2025lingotrip} demonstrate that LLMs can interpret spatial cues and infer user intentions from trajectory-like sequences. Prior methods, however, do not compress noisy long-term usage logs nor model fine-grained app and traffic semantics, which motivates our hierarchical LLM-guided framework.

%% file: Conclusion.tex
\section{Conclusion}
\label{sec:conclusion}
In this paper, we propose LLMApp, a hierarchical prediction framework that leverages large language models to predict mobile app usage. 
By decomposing prediction into category-level intent inference and fine-grained app and traffic estimation, LLMApp effectively captures the hierarchical structure underlying human app usage behavior. The proposed prompt compressor distills long-term records into concise and informative prompts, while the context interpreter embeds diverse contextual factors into a unified semantic representation space. These components enable the LLM to reason over spatiotemporal trajectories, user habits, and semantic app relationships in a unified manner.
Comprehensive experiments across two real-world city-scale datasets show that LLMApp consistently outperforms baselines. Beyond accuracy, our analyses demonstrate that LLMApp learns more coherent spatial clusters, provides stable performance across hyperparameters, and exhibits strong cross-city robustness. The results highlight that LLMs, when combined with carefully designed context encoding and hierarchical reasoning, can serve as potent predictors for complex human behavioral sequences.
In future work, the framework can be extended toward broader mobility-behavior modeling tasks, multi-agent prediction, and real-time adaptive mobile service optimization. LLMApp shows that semantic reasoning and cross-modal unification enabled by LLMs open promising avenues for modeling human–device interaction at scale.